\documentclass[unsrt]{article}
\usepackage[a4paper,total={5.5in, 8.5in}]{geometry}
\usepackage{framed,multirow}
\usepackage{url}
\usepackage{xcolor}
\usepackage{hyperref}
\usepackage{graphicx}
\usepackage{tabularx}
\usepackage{amssymb}
\usepackage{bbold}
\usepackage{rotating}
\usepackage{array}
\usepackage{cuted}
\usepackage{lipsum}     
\usepackage{caption}
\usepackage{amsmath}
\usepackage{mathtools}
\usepackage[caption=false]{subfig}
\usepackage{amsfonts}
\usepackage{enumerate}
\usepackage{algorithmicx}
\usepackage{algorithm2e}
\usepackage{algpseudocode}
\usepackage{fancyhdr}
\usepackage{textcomp}
\usepackage{gensymb}
\usepackage{bm}
\usepackage{algcompatible}

\usepackage{algpascal}

\newcommand{\dd}{
\mathop{}\mathopen{}\mathrm{d}
}
\usepackage{lineno}

\title{Linear Anchored Gaussian Mixture Model for Location and Width Computations of Objects in Thick Line Shape}

\author{Nafaa Nacereddine*\textonesuperior, Aicha B. Goumeidane\textonesuperior,  Djemel Ziou\texttwosuperior \\ 
\small\textonesuperior Research Center in Industrial Technologies CRTI, P.O.Box 64, Ch\'{e}raga, Algiers 16014, Algeria\\
\small\texttwosuperior DMI, Universit\'{e} de Sherbrooke, Qu\'{e}bec, QC J1K 2R1, Canada
\thanks{N.Nacereddine is the corresponding author, email: n.nacereddine@crti.dz}
}

\begin{document}
\maketitle
\begin{abstract}
Accurate detection of the centerline of a thick linear structure and good estimation of its thickness are challenging topics in many real-world applications such X-ray imaging, remote sensing and lane marking detection in road traffic. Model-based approaches using Hough and Radon transforms are often used but, are not recommended for thick line detection, whereas methods based on image derivatives need further step-by-step processing making their efficiency dependent on each step outcome. In this paper, a novel paradigm to better detect thick linear objects is presented, where the 3D image gray level representation is considered as a finite mixture model of a statistical distribution, called linear anchored Gaussian distribution and parametrized by a scale factor to describe the structure thickness and radius and angle parameters to localize the structure centerline. Expectation-Maximization algorithm (Algo1) using the original image as input data is used to estimate the model parameters. To rid the data of irrelevant information brought by nonuniform and noisy background, a modified EM algorithm (Algo2) is detailed. 
In Experiments, the proposed algorithms show promising results on real-world images and synthetic images corrupted by blur and noise, where Algo2, using Hessian-based angle initialization, outperforms Algo1 and Algo2 with random angle initialization, in terms of running time and structure location and thickness computation accuracy.
\end{abstract}


\section{Introduction}
\label{intro}
The detection of thin linear structures in an image is a very important task in computer vision, where, nowadays, increasingly sophisticated methods are developed to recover such structures in the most precise way despite the challenging characteristics that may affect the input images such as bad contrast, noise, occlusion, artifacts, complex background, etc. This quest of precision is motivated by the requirements of involved sensitive applications of which a lot of specific examples could be named like, road extraction from remotely sensed imagery \cite{Wang2016}, strip scanning in radio astronomy, detection of linear defect indications in weld radiographic testing \cite{Nacereddine2019}, detection of road markings in road traffic scenes \cite{Chen2020}, pleural line detection in lung ultrasound imaging \cite{McDermott2021}, detection and classification of linear structures in mammographic images \cite{Zwiggelaar2004}. To deal with the above-posed problem, many approaches can be found in literature.

For thick line detection in images, some methods based on image function derivatives used alone or in combination with geometrical computations are applied \cite{Miao2014,Even2019}. However, the drawbacks of these methods reside in their needs to additional processing to deal with the effects of noise and blur, in addition of the fact that they are considered as step-by-step methods which make their outcomes quality dependent on the quality of each processed step. 

Recently, number of works are dedicated to thick linear structures detection in images using deep learning such as road extraction in remote sensing \cite{Xu2018} and lane detection in traffic road marking \cite{Tian2018,Tang2021}. However, these techniques are faced to some  challenges: (1) a lack of generalization since supervised learning requires a large amount of annotated data, and labeling data is a boring and costly task, and (2) an expensive computation since Convolutional Neural Network (CNN) architectures are involving millions of parameters.

About the model-based approaches, the Hough transform and the Radon transform are always recommended for detection of thin lines but fail, unless adding tricks, to detect the thick line because, for the latter, the diagonal is detected instead of the centerline \cite{Xu14}. However, a recent integral transform-based method, called scale space Radon transform (SSRT) \cite{Ziou2021}, has been shown to be more general since its parameterization allows to handle the structure thickness during detection, permitting to recover efficiently the centerline of a thick linear object, even in the presence of noise and blur. Nevertheless, this method has inherited, from the Radon transform, its sensitivity to background complexity when the linear structure is quite small compared to the image dimensions. In addition to the work citepd in \cite{Ziou2021}, a part of state-of-the-art dedicated to thick line detection and extraction can be found in the same reference.        
In this paper, for the abovementioned problematic, another model-based approach is investigated, namely, finite mixture model where, the input image, composed of a number of thick linear structures and converted to a grayscale image, is modeled as a mixture of parameterized statistical distribution. 

By observing the gray level relief of an image containing thick lines, the first motivation behind the choice of this approach is the ability to find a statistical distribution that faithfully approximates all the parameters governing a thick line in an image while, the utilization of a mixture model is the consequence of the possible presence of several of these structures in a single image. It is supposed that, on a finite dimensions image, the normalized 3D gray level representation of a thick line follows a linear anchored Gaussian probability density function of which the analytical expression will be detailed in the next section. We aimed in this work to estimate, for each linear structure, the parameters relative to its space occupancy rate, its thickness and the parameters related to its centerline location i.e. radius and orientation angle. Such parameters estimation is achieved by using the Expectation-Maximization (EM) algorithm \cite{Dempster1977, McLachlan2000}.

In the major part of real world imagery applications, the images embedding the linear structures we aim to detect are very noisy and, sometimes, present very complex background. That is why, in this paper, to deal with this problem, an adaptive selection of the image data based on dynamic background subtraction, required in the likelihood function computation, will be proposed.     
 
The remaining of the paper is organized as follows. Section 2 will detail the new proposed linear anchored Gaussian distribution. The Section 3 will be devoted to the estimation of the finite linear anchored Gaussian mixture model parameters through EM algorithm. Synthetic and real images with noise and complex backgrounds are used in Experiments in Section 4, where a new data selection paradigm during parameter estimation is presented and tested. Concluding remarks and perspectives are drawn in Section 5. Two Appendix subsections are provided and the end of this paper, before the references.

\section{Linear anchored Gaussian distribution}
\begin{figure}[tbh!]
\centering
\includegraphics[scale=0.4]{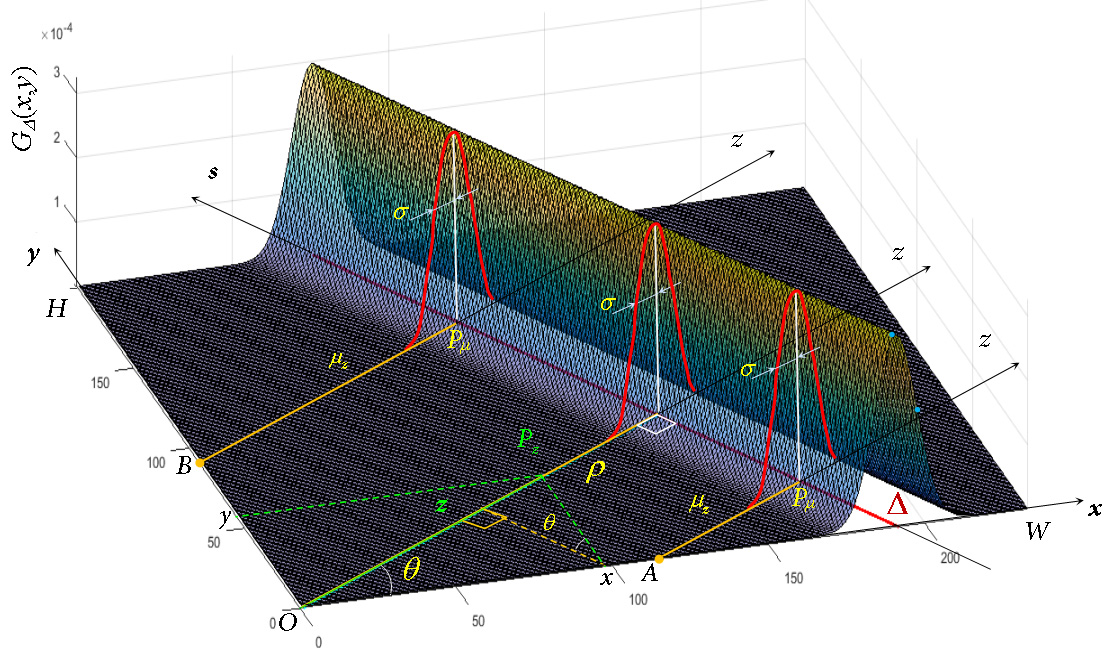}
\caption{Linear anchored Gaussian function with $\sigma=5.8$ for the parameterized line $(\Delta)$ with $\rho=166$ and $\theta=\pi/6$} \label{l_Gauss}
\end{figure}

To introduce the formulation of a linear anchored Gaussian distribution (LAGD), let us give the example of the presence, in an image of width $W$ and height $H$, of a thick linear object on an uniform background. The 3D gray level representation of the linear structure could be seen as a rectangular parallelepiped, delimited by the image domain $\mathcal{D}$, of which the longest symmetry axis of its basis is described by a parameterized equation $(\Delta): x\cos\theta+y\sin\theta-\rho=0$ where $\rho$ and $\theta$ are the radius and the angle parameters representing, respectively, the distance from the $x-y$ plane origin to the line $(\Delta)$ and the angle formed between the latter and the $x$-axis, as described in Fig.~\ref{l_Gauss}. 

The 1D profile taken orthogonal to $\Delta$, as seen in Fig.~\ref{l_Gauss}, consists in an univariate Gaussian distribution of random variable $\mathcal{Z}$, noted $\mathcal{N}(\mu_z,\sigma)$, of which the $pdf$ is expressed as $g_z(\bm{z}|\mu_z,\sigma)=\frac{1}{\sqrt{2\pi}\sigma}\exp{\frac{-(z-\mu_z)^2}{2\sigma^2}}$. In the $x-y$ plane, the coordinates of the point $P_z$ representing the data element $z$ are $(x,y)$ while the point $P_\mu \ (P_\mu\in\Delta)$, representing the mean parameter $\mu_z$ has as coordinates $(\mu_x,\mu_y)$ where, $\mu_x\cos\theta+\mu_y\sin\theta-\rho=0$. We recall that all the $pdf$s of the Gaussians stacked perpendicularly to $\Delta$ have the same standard deviation $\sigma$.

By sliding the $z$-axis perpendicularly to $\Delta$ to include all the image data while replacing $z$ and $\mu_z$ by the computed values in each of the configurations depicted in \ref{apdx_A}, we obtain a linear anchored Gaussian function $G_{\Delta}$ expressed in the $x-y$ plane as 
\setlength{\abovedisplayskip}{6pt}
\setlength{\belowdisplayskip}{6pt}
\begin{equation}
G_{\Delta}(x,y|\rho,\theta,\sigma)=\frac{1}{\sqrt{2\pi}\sigma}\exp{\frac{-(x\cos\theta+y\sin\theta-\rho)^2}{2\sigma^2}}
\label{G_eq}
\end{equation}

where, $\rho$ and $\theta$ are already defined above and illustrated in Fig.~\ref{l_Gauss}. Still about Eq.~\ref{G_eq}, the parameter $\sigma$ is the standard deviation which is proportional to the thickness of the linear structure to estimate and which is the same for all the univariate Gaussians embedded by the $z$-axis, stacked perpendicularly to $\Delta$ and whose the mean parameters coordinates verify the canonical equation of $\Delta$. Let $\bm{\phi}=(\rho,\theta,\sigma)$ denoting the parameters vector of $G_{\Delta}$ function. 

The grayscale image, represented by $G_{\Delta}$ function, is defined on a rectangular domain $\mathcal{D}=W\times H$ included in the plan $\mathbb{R}^2$ and takes its values in the grayscale range of the image. 

Let us denote $g_{\Delta}$ the normalized linear anchored Gaussian function, expressed as
\setlength{\abovedisplayskip}{6pt}
\setlength{\belowdisplayskip}{6pt}
\begin{equation}
g_{\Delta}(x,y|\bm{\phi}) = 
     \begin{cases}
        \frac{G_\Delta(x,y|\bm{\phi})}{\iint_\mathcal{D} G_\Delta(x,y|\bm{\phi})} & x,y\in\mathcal{D}\\
        \small 0  & \small\text{elsewhere}
     \end{cases}
\label{g_delt_eq}
\end{equation} 

Since $g_{\Delta}(x,y|\rho,\theta,\sigma)\geq 0$ for $x,y\in\mathcal{D}$ and $\iint_\mathcal{D}g_{\Delta}(x,y|\bm{\phi})dxdy=1$, the function $g_{\Delta}$ can be processed as a $pdf$ of the continuous 3D representation of a linear structure in a grayscale image. It is worthy to note that the amount $\iint_\mathcal{D} G_\Delta(x,y|\bm{\phi})$ is nothing but the volume under the $G_{\Delta}$ function, noted $V^{(\Delta)}$.

\section{Image intensities as a mixture of LAGDs} 

\begin{figure}[tbh!]
\centering
\includegraphics[scale=0.4]{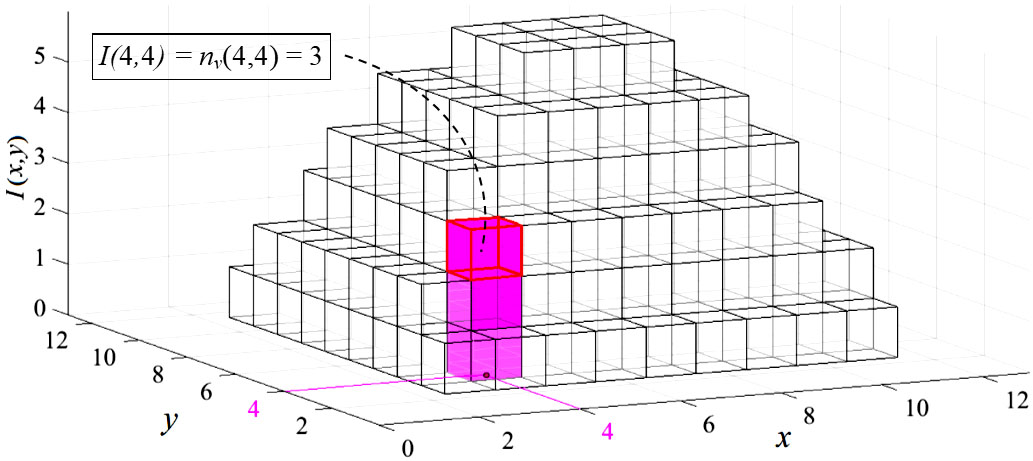}
\caption{An example of unitary volume illustration in the image relief related to the realizations of $\bm{v}$}
\label{unit_vol}
\end{figure}

Let $I$ be a grayscale image of size $N=W\times H$ pixels of which gray values are defined in $\{0,1,\dots,L-1\}$. As illustrated in Fig.~\ref{unit_vol}, the 3D gray level representation of the image can be considered as a 3D space dataset or a volume, called $\bm{v}$, consisting in $N_v$ unitary volumes. It results that $N_v= \sum_{i=1}^N g_i=\sum_{x=1}^W\sum_{y=1}^H n_v(x,y)$ where, $g_i$ is the gray value of the pixel $i$ among the whole $N$ pixels and $n_v(x,y)$ denotes the number of volume units stacked on the pixel of coordinates $(x,y)$ which is nothing but the pixel gray value $I(x,y)$.
Our key idea is to consider the volume $\bm{v}$ as a collection of observations of a random variable $\mathcal{V}$ following a distribution having $f$ as $pdf$, i.e. $\bm{v} = \{v_i|i=1,\dots,N_v\}$ being the set of realizations of $\mathcal{V}$. 

Since, in this work, it is question to detect linear structures or objects in an image, the gray levels relief of the latter could be assumed to be fitted by a linear anchored Gaussian mixture model of which the $pdf$ is expressed as  
\begin{equation}
f(\bm{v}|\bm{\Phi}) = \sum\limits_{m=1}^{M}\pi_{m}g_{\Delta}(\bm{v}|\bm{\phi}_m)
\label{pdfmix}
\end{equation}
where, $M$ denotes the number of linear structures contained in the image and $\bm{\Phi} = (\pi_1,\dots,\pi_M,\\\bm{\phi_1},\dots,\bm{\phi}_M)$ is the mixture model parameter vector with $\pi_m\mid\sum_{m=1}^M\pi_m=1$ and $\bm{\phi}_m=(\rho_m,\theta_m,\sigma_m)$ are, respectively, the proportion and the distribution parameters of the $m$th mixture component.

\begin{figure}[tbh!]
\centering
\includegraphics[scale=0.6]{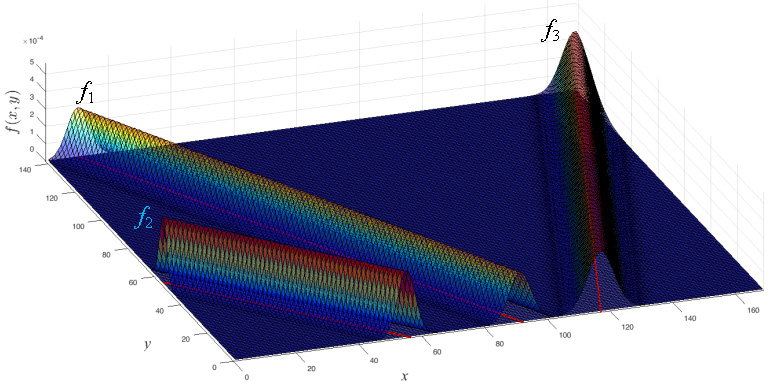}
\caption{An example of a mixture of linear anchored Gaussian distributions}
\label{l_GaussMix}
\end{figure}

Let us give in Fig.~\ref{l_GaussMix} an example of graphics representing a mixture of three ($M$=3) linear anchored Gaussian \textit{pdf}s defined by $f(\bm{v}|\bm{\Phi}) =\sum_{m=1}^3\pi_{m}g_{\Delta}(\bm{v}|\bm{\phi}_m)= \sum_{m=1}^3 f_m$ where $\bm{\pi}=[0.35 \ 0.15 \ 0.5]$, $\bm{\phi}_1=[80\ \pi/6 \; 3]$, $\bm{\phi}_2=[40 \ \pi/4 \; 2]$ and $\bm{\phi}_3=[110 \ -\pi/9 \; 4]$.

The likelihood function, for the $pdf$ mixture given in Eq.~\ref{pdfmix}, is expressed as  
\begin{equation}
\mathcal{L} = \prod\limits_{i=1}^{N_v}\sum\limits_{m=1}^{M}\pi_{m}g_{\Delta}(v_i|\bm{\phi}_m)
\label{likfunc}
\end{equation}

Recall that each realization $v_i$ of the random variable $\mathcal{V}$ corresponds to volume unit located at $x-y$ coordinates of the image (See Fig.~\ref{unit_vol}). Eq.~\ref{likfunc} can be detailed as follows:
\setlength{\abovedisplayskip}{6pt}
\setlength{\belowdisplayskip}{6pt}
\small
\begin{align}
\mathcal{L} = \underbrace{\sum\mathop{}_{\mkern-5mu m}\pi_{m}g_{\Delta}(1,1|\bm{\phi}_m)\times\dots\times\sum\mathop{}_{\mkern-5mu m}\pi_{m}g_{\Delta}(1,1|\bm{\phi}_m)}\limits_{n_v(1,1) \ \text{times}}\times\nonumber\\
 \underbrace{\sum\mathop{}_{\mkern-5mu m}\pi_{m}g_{\Delta}(1,2|\bm{\phi}_m)\times\dots\times\sum\mathop{}_{\mkern-5mu m}\pi_{m}g_{\Delta}(1,2|\bm{\phi}_m)}\limits_{n_v(1,2)\ \text{times}}\times\dots\nonumber\\
 \times\underbrace{\sum\mathop{}_{\mkern-5mu m}\pi_{m}g_{\Delta}(W,H|\bm{\phi}_m)\times\dots\times\sum\mathop{}_{\mkern-5mu m}\pi_{m}g_{\Delta}(W,H|\bm{\phi}_m)}\limits_{n_v(W,H) \ \text{times}}
\label{likfunc1}
\end{align}
\normalsize

This equation arises from the fact that, at each position $(x,y)$, the observation is the set of $n_v(x,y)$ unitary volumes stacked at $(x,y)$. 
Then, by regrouping the observations $v_i$ together according to their $x-y$ coordinates, the likelihood function expression can be developed as follows
\setlength{\abovedisplayskip}{6pt}
\setlength{\belowdisplayskip}{6pt}
\small
\begin{align}\nonumber
\mathcal{L}
&=\prod\limits^{n_v(1,1)}\sum\limits_m\pi_{m}g_{\Delta}(1,1|\bm{\phi}_m)\times\prod\limits^{n_v(1,2)}\sum\limits_m\pi_{m}g_{\Delta}(1,2|\bm{\phi}_m)\times\dots\prod\limits^{n_v(1,H)}\sum\limits_m\pi_{m}g_{\Delta}(1,H|\bm{\phi}_m)\\\nonumber
&\times\prod\limits^{n_v(2,1)}\sum\limits_m\pi_{m}g_{\Delta}(2,1|\bm{\phi}_m)\times\dots
\prod\limits^{n_v(W,1)}\sum\limits_m\pi_{m}g_{\Delta}(W,1|\bm{\phi}_m)\times\dots\prod\limits^{n_v(W,H)}\sum\limits_m\pi_{m}g_{\Delta}(W,H|\bm{\phi}_m)\\
&=\prod\limits_{x=1}^W\prod\limits_{y=1}^H\prod\limits^{n_v(x,y)}\sum\limits_m\pi_{m}g_{\Delta}(x,y|\bm{\phi}_m)
=\prod\limits_{x=1}^{W}\prod\limits_{y=1}^{H}\bigg[\sum\limits_m\pi_{m}g_{\Delta}(x,y|\bm{\phi}_m)\bigg]^{n_v(x,y)}
\label{likfunc2}
\end{align}
\normalsize
Knowing that $n_v(x,y)= I(x,y)$, the log-likelihood function is then deduced as 
\small
\begin{equation}
\log\mathcal{L} = \sum\limits_{x=1}^{W}\sum\limits_{y=1}^{H}I(x,y)\log\sum\limits_{m=1}^{M}\pi_{m}g_{\Delta}(x,y|\bm{\phi}_m)
 = N_v\sum\limits_{x=1}^{W}\sum\limits_{y=1}^{H}h(x,y)\log\sum\limits_{m=1}^{M}\pi_{m}g_{\Delta}(x,y|\bm{\phi}_m)
\label{likfunc4}
\end{equation}
\normalsize
where $h$ is the normalized image which approximate the image $pdf$ expressed as 
\setlength{\abovedisplayskip}{6pt}
\setlength{\belowdisplayskip}{6pt}
\begin{equation}
h(x,y)=\frac{I(x,y)}{\sum_{x=1}^{W}\sum_{y=1}^{H}I(x,y)}=\frac{I(x,y)}{N_v}
\end{equation}

The log-likelihood function as given above is difficult to optimize because it contains the log of the sum \cite{Bilmes1997}. If we consider $\mathcal{V}$ as a partial observation of data, one can introduce \textit{M}-dimensional random variable $\mathcal{Z}$ representing the unobserved data $\bm{z}=\{z_{xy}|1\leq x\leq W, 1\leq y\leq H\}$ whose values represent the labels of the incomplete data $\bm{v}$. Then, by introducing the realization $\bm{z}_{xy} = (z_{xy1},\dots,z_{xyM})$,  the complete data $\mathcal{Y} = (\mathcal{V},\mathcal{Z})$ is formed, where $z_{xym}=1$ and $z_{xyk}=0$ for $k\neq m$
indicate that gray level $I(x,y)$ is produced by the $m$th component of the mixture. The random variable $\mathcal{Z}$ is generally supposed to be a multinomial random variable parameterized by $\bm{\pi} = (\pi_1,\dots,\pi_M)$. Under these assumptions, the resulting complete log-likelihood function is given by
\setlength{\abovedisplayskip}{6pt}
\setlength{\belowdisplayskip}{6pt}
\begin{equation}
\mathcal{L}_c(\Phi) = N_v\sum\limits_{x=1}^{W}\sum\limits_{y=1}^{H}\sum\limits_{m=1}^{M}z_{xym}h(x,y)\log\left(\pi_{m}g_{\Delta}(x,y|\bm{\phi}_m)\right)
\end{equation}

The EM algorithm \cite{Dempster1977,McLachlan2000} is an iterative algorithm alternating, at iteration $t$, between performing an Expectation step (E-step), which computes the log-likelihood expectation, noted  $Q$, evaluated using the current estimate for the parameters $\bm{\Phi}{(t)}$, and a Maximization step (M-step), which  comes to maximize the $Q-$function to estimate the new model parameter $\bm{\Phi}^{(t+1)}$.

\subsection{E-step}

E-step computes, at iteration $t$, the expectation of the log-likelihood function $\mathcal{L}_{c}(\bm{\Phi^{(t)}})$ called the $Q-$function and denoted $Q(\bm{\Phi}|\bm{\Phi}^{(t)})$, i.e., 
\setlength{\abovedisplayskip}{6pt}
\setlength{\belowdisplayskip}{6pt}
\begin{equation}
Q(\bm{\Phi}|\bm{\Phi}^{(t)})=\mathbb{E}_{\mathcal{Z}|\mathcal{V},\bm{\Phi}^{(t)}}\left[\mathcal{L}_{c}(\bm{\Phi})\right]
\label{Q_eq1}
\end{equation}

Since the only random part of Eq.~\ref{Q_eq1} is $z_{xym}$ and the remainder is deterministic, then 
\setlength{\abovedisplayskip}{6pt}
\setlength{\belowdisplayskip}{6pt}
\begin{equation}
Q(\bm{\Phi}|\bm{\Phi}^{(t)})= 
N_v\sum\limits_{x=1}^{W}\sum\limits_{y=1}^{H}\sum\limits_{m=1}^{M}\mathbb{E}_{\mathcal{Z}}\left[z_{xym}\right]h_{xy}\log\left(\pi_{m}g_{\Delta}(x,y|\bm{\phi}_m)\right)
\label{Q_eq2}
\end{equation}
where, $h_{xy}$ is the contracted notation of $h(x,y)$ and
\begin{equation}
\mathbb{E}\left[z_{xym}\right] = 0\!\times\!Pr(z_{xym}\!=\!0|x,y)\!+\!1\!\times\!Pr(z_{xym}\!=\!1|x,y)=Pr(z_{xym}\!=\!1|x,y) 
\end{equation}

By using the Bayes' theorem, we obtain
\begin{equation}
Pr(z_{xym}\!=\!1|x,y)=\frac{Pr(z_{xym}\!=\!1)Pr(x,y|z_{xym}\!=\!1)}{Pr(x,y)}\nonumber
\end{equation}

At iteration $t$, we have $Pr(z_{xym}\!=\!1)\!=\!\pi_m^{(t)}$ and $Pr(x,y)=\sum_{m=1}^M \pi_m^{(t)}g_{\Delta}\big(x,y|\bm{\phi}_m^{(t)}\big)$, 
then $Pr(x,y|z_{xym}\!=\!1)=g_{\Delta}\big(x,y|\bm{\phi}_m^{(t)}\big)$. It results,
\begin{equation}
Pr(z_{xym}\!=\!1|x,y)=z_{xym}^{(t)}=\frac{\pi_m^{(t)}g_{\Delta}\big(x,y|\bm{\phi}_m^{(t)}\big)}{\sum_{l=1}^M\pi_l^{(t)}g_{\Delta}\big(x,y|\bm{\phi}_l^{(t)}\big)}
\label{Estep}
\end{equation}
where, $z_{xym}^{(t)}$ are the posterior probabilities, computed at iteration $t$, using the current estimates of the model parameters $\big(\pi_m^{(t)},\phi_m^{(t)}\big)=\bm\Phi_m^{(t)}$. 

The expression of the $m$-component of the linear anchored Gaussian mixture, at iteration $t$, is given by
\setlength{\abovedisplayskip}{4pt}
\setlength{\belowdisplayskip}{4pt}
\small
\begin{equation}
g_{\Delta}\big(x,y|\bm{\phi}_m^{(t)}\big)=\frac{1}{V^{(\Delta)}_m\sigma_m^{(t)}\sqrt{2\pi}}
\exp{\frac{-\big(x\cos\theta_m^{(t)}+y\sin\theta_m^{(t)}-\rho_m^{(t)}\big)^2}{2\sigma_m^{2(t)}}}
\label{g_eq_iter}
\end{equation}
\normalsize
where $V^{(\Delta)}_m$ is the volume under the $m$th linear anchored Gaussian, as mentioned, in \S 2. 

By developing the expression of $Q$ given in Eq.~\ref{Q_eq2}, we obtain
\setlength{\abovedisplayskip}{6pt}
\setlength{\belowdisplayskip}{6pt}
\begin{multline}
Q(\bm{\Phi}|\bm{\Phi}^{(t)})=N_v\sum\limits_{x=1}^{W}\sum\limits_{y=1}^{H}\sum\limits_{m=1}^{M}z_{xym}^{(t)}h_{xy}\left[\log\pi_{m}-\log(V_m^{(\Delta)}\sigma_m\sqrt{2\pi})\right.\\
\left.-\frac{(x\cos\theta_m+y\sin\theta_m-\rho_m)^2}{2\sigma_m^2}\right]
\label{Q_it_eq}
\end{multline}

\subsection{M-step}
In M-Step, the parameter estimates $\widehat{\bm{\Phi}}=(\widehat{\pi}_m,\widehat{\rho}_m,\widehat{\theta}_m,\widehat{\sigma}_m)_{m=1,\dots,M}$ are computed by updating the parameters according to $\bm{\Phi}^{(t+1)}=\arg\max\limits_{\Phi} Q(\bm{\Phi}|\bm{\Phi}^{(t)})$
until the convergence is achieved. The parameter estimates, at iteration $t\!+\!1$, are derived in the following paragraphs where, for the sake of simplicity, $Q(\bm{\Phi}|\bm{\Phi}^{(t)})$ is noted $Q$ and the iteration indexing is added only to the final expression of each parameter. 

\subsubsection{Mode proportion in the mixture $\pi_m$}
Because of the constraint $\sum_{m=1}^M\pi_m=1$ on the mixture modes, the Lagrange multiplier $\lambda$ is introduced and the $\pi_m$ estimate is derived as follows:
\begin{align}\nonumber
&\frac{\partial}{\partial\pi_m}\left[N_v\sum\limits_{x=1}^{W}\sum\limits_{y=1}^{H}\sum\limits_{m=1}^{M}z_{xym}^{(t)}h_{xy}
\log\pi_{m}+\lambda\left(\sum_{m=1}^M\pi_m-1\right)\right]\\
&=\frac{N_v}{\pi_m}\left[\sum\limits_{x=1}^{W}\sum\limits_{y=1}^{H}z_{xym}^{(t)}h_{xy}+\lambda\right]=0
\end{align}
Then, $\pi_m=N_v\sum\limits_{x=1}^{W}\sum\limits_{y=1}^{H}z_{xym}^{(t)}h_{xy}/(-\lambda)$   
and $\sum\limits_{m=1}^M\pi_m=N_v\sum\limits_{x=1}^{W}\sum\limits_{y=1}^{H}\sum\limits_{m=1}^Mz_{xym}^{(t)}h_{xy}/(-\lambda)=1$
Since $\sum\limits_{x=1}^{W}\sum\limits_{y=1}^{H}h_{xy}=1$ and $\sum\limits_{m=1}^M z_{xym}^{(t)}=1$ then $-\lambda=N_v\sum\limits_{x=1}^{W}\sum\limits_{y=1}^{H}\sum\limits_{m=1}^{M}z_{xym}^{(t)}h_{xy}=N_v$. Finally, we have
\begin{equation}
\pi_m^{(t+1)}=\sum\limits_{x=1}^{W}\sum\limits_{y=1}^{H}z_{xym}^{(t)}h_{xy}
\label{pi_est}
\end{equation}

\subsubsection{Radius parameter $\rho_m$}
By setting to zero the derivative of $Q$ w.r.t. $\rho_m$, we obtain 
\setlength{\abovedisplayskip}{4pt}
\setlength{\belowdisplayskip}{4pt}
\begin{equation}
\frac{\partial Q}{\partial\rho_m} = -\frac{N_v}{2\sigma_m^2}\sum\limits_{x=1}^{W}\sum\limits_{y=1}^{H}z_{xym}^{(t)}h_{xy} 2(-1)(x\cos\theta_m+y\sin\theta_m-\rho_m)=0
\end{equation}
yielding to
\setlength{\abovedisplayskip}{4pt}
\setlength{\belowdisplayskip}{4pt}
\begin{align}\nonumber
\rho_m^{(t+1)} = \frac{\sum_{x=1}^{W}\sum_{y=1}^{H}z_{xym}^{(t)}h_{xy}(x\cos\theta_m^{(t)}+y\sin\theta_m^{(t)})}{\sum_{x=1}^{W}\sum_{y=1}^{H}z_{xym}^{(t)}h_{xy}}\\
 = \frac{1}{\pi_m^{(t)}}\sum\limits_{x=1}^{W}\sum\limits_{y=1}^{H}z_{xym}^{(t)}
h_{xy}(x\cos\theta_m^{(t)}+y\sin\theta_m^{(t)})
\label{rho_est}
\end{align}

\subsubsection{Scale parameter $\sigma_m$}
By setting to zero the derivative of $Q$ w.r.t. $\sigma_m$, we have 
\setlength{\abovedisplayskip}{4pt}
\setlength{\belowdisplayskip}{4pt}
\begin{multline}
\frac{\partial Q}{\partial\sigma_m} = -\frac{N_v}{\sigma_m}\sum\limits_{x=1}^{W}\sum\limits_{y=1}^{H}z_{xym}^{(t)}h_{xy}\\
+N_v\sum\limits_{x=1}^{W}\sum\limits_{y=1}^{H}z_{xym}^{(t)}h_{xy}\frac{4\sigma_m(x\cos\theta_m+y\sin\theta_m-\rho_m)^2}{4\sigma_m\sigma_m^3}=0 
\end{multline}

The scale parameter $\sigma_m$ is then obtained as 
\setlength{\abovedisplayskip}{4pt}
\setlength{\belowdisplayskip}{4pt}
\begin{align}\nonumber
\sigma_m^{(t+1)}&=\left[\frac{\sum\limits_{x=1}^{W}\sum\limits_{y=1}^{H}z_{xym}^{(t)}h_{xy}\left(x\cos\theta_m^{(t)}\!+\!y\sin\theta_m^{(t)}\!-\!\rho_m^{(t)}\right)^2}{\sum_{x=1}^{W}\sum_{y=1}^{H}z_{xym}^{(t)}h_{xy}}\right]^{\frac{1}{2}}\\
&=\left[\frac{1}{\pi_m^{(t)}}\sum\limits_{x=1}^{W}\sum\limits_{y=1}^{H}z_{xym}^{(t)}h_{xy}\left(x\cos\theta_m^{(t)}\!+\!y\sin\theta_m^{(t)}\!-\!\rho_m^{(t)}\right)^2\right]^{\frac{1}{2}}
\label{sig_est}
\end{align}

In an ideal case, if the gray level relief of the linear structure has a perfect rectangular parallelepiped with thickness $w$, then we can deduce the relationship between the standard deviation $\sigma_m$ estimating a thick line $m$ and the thickness $w_m$ of the latter as (See the proof in \ref{apdx_B})
\setlength{\abovedisplayskip}{4pt}
\setlength{\belowdisplayskip}{4pt}
\begin{equation}
2\sigma_m = w_m/\sqrt{3}
\label{final_sig_wid_rel}
\end{equation}

\subsubsection{Angle parameter $\theta_m$}
By setting to zero the derivative of $Q$ w.r.t. $\theta_m$, we obtain 
\setlength{\abovedisplayskip}{4pt}
\setlength{\belowdisplayskip}{4pt}
\begin{equation}
\frac{\partial Q}{\partial\theta_m} = -\frac{N_v}{2\sigma_m^2}\sum\limits_{x=1}^{W}\sum\limits_{y=1}^{H}z_{xym}^{(t)}h_{xy}
(-x\sin\theta_m+y\cos\theta_m)(x\cos\theta_m+y\sin\theta_m-\rho_m)=0
\label{theta_opt1}
\end{equation}

Let note the amounts \small$\sum\limits_x\sum\limits_yz_{xym}^{(t)}h_{xy}x$, $\sum\limits_x\sum\limits_yz_{xym}^{(t)}h_{xy}y$, $\sum\limits_x\sum\limits_yz_{xym}^{(t)}h_{xy}xy$, $\sum\limits_x\sum\limits_yz_{xym}^{(t)}h_{xy}x^2$\normalsize \ and \small$\sum\limits_x\sum\limits_yz_{xym}^{(t)}h_{xy}y^2$ by $m_x, m_y, m_{xy}, m_{x^2}, m_{y^2}$\normalsize, respectively.

If we replace $\tan\theta_m$ by $u_m$ then the couple $(\sin\theta_m,\cos\theta_m)$ will take the values $\frac{1}{\sqrt{1+u_m^2}}(u_m,1)$ and $-\frac{1}{\sqrt{1+u_m^2}}(u_m,1)$ leading, from Eq.~\ref{theta_opt1}, to the following respective equations   
\setlength{\abovedisplayskip}{4pt}
\setlength{\belowdisplayskip}{4pt}
\begin{equation}
(m_{x^2}-m_{y^2})u_m - m_{xy}(1-u_m^2)-\rho_m(m_xu_m-m_y)\sqrt{1+u_m^2}=0
\label{theta_opt2}
\end{equation}
\begin{equation}
(m_{x^2}-m_{y^2})u_m - m_{xy}(1-u_m^2)+\rho_m(m_xu_m-m_y)\sqrt{1+u_m^2}=0
\label{theta_opt3}
\end{equation}

Let us denote the polynoms in the left sides of the equations Eq.~\ref{theta_opt2} and Eq.~\ref{theta_opt3} by $P_1(\theta_m)$ and $P_2(\theta_m)$, respectively. The optimized values of the angle parameter will belong to the roots of the four-order polynomial equation, called also quartic equation, formed, from the the product of $P_1(\theta_m)$ and $P_2(\theta_m)$, as given in Eq.~\ref{theta_opt4}
\setlength{\abovedisplayskip}{4pt}
\setlength{\belowdisplayskip}{4pt}
\begin{align}\nonumber
P(\theta_m)=P_1(\theta_m)\times P_2(\theta_m)&\\\nonumber
=(m_{x^2}\!-\!m_{y^2})^2u_m^2\!-\!2m_{xy}(m_{x^2}\!-\!m_{y^2})(u_m\!-\!u_m^3)\!+\!m_{xy}^2(1\!-\!2u_m^2\!+\!u_m^4)&\\\nonumber
-\rho_m^2(m_x^2u_m^2\!-\!2m_xm_yu_m\!+\!m_y^2)(1\!+\!u_m^2)&\\
=au_m^{4(t+1)}\!+\!bu_m^{3(t+1)}\!+\!cu_m^{2(t+1)}\!+\!du_m^{(t+1)}\!+\!e=0& 
\label{theta_opt4}
\end{align}
where, 
\begin{equation}\nonumber
\begin{cases}
a = m_{xy}^{2(t)}\!-\!\rho_m^{2(t)}m_x^{2(t)}\\
b = 2m_{xy}^{(t)}(m_{x^2}^{(t)}\!-\!m_{y^2}^{(t)})\!+\!2\rho_m^{2(t)}m_x^{(t)}m_y^{(t)}\\
c = \left(m_{x^2}^{(t)}\!-\!m_{y^2}^{(t)}\right)^2\!-\!2m_{xy}^{2(t)}\!-\!\rho_m^{2(t)}(m_x^{2(t)}\!+\!m_y^{2(t)})\\
d = 2\rho_m^{2(t)}m_x^{(t)}m_y^{(t)}\!-\!2m_{xy}^{(t)}(m_{x^2}^{(t)}\!-\!m_{y^2}^{(t)})\\
e = m_{xy}^{2(t)}\!-\!\rho_m^{2(t)}m_y^{2(t)}\\
\end{cases}\,
\end{equation}

The possible cases of the nature of the four roots of $P_{\theta_m}$, given in Eq.~\ref{theta_opt4}, are as follows: distinct, double, multiple, real, complex conjugate.  
From the roots quoted before, we retain only the real solutions, i.e. $\{\theta_m^{(t+1)}=\arctan u_m^{(t+1)}| u_m^{(t+1)}\in\mathbb{R}\}$; afterward, from the latter, we keep only the solution that maximizes the function $Q$. 

\subsection{Algorithm}
For the proposed mixture model-based linear structure detection in images, the various operations followed by the EM algorithm are summarized in Algo 1.

\begin{algorithm}
\caption{EM algorithm for linear anchored Gaussian mixture model}
\KwIn{$h$: Normalized input grayscale image\\
\hspace{1cm} $M$: Number of mixture components\\
\hspace{1cm} Initial mixture parameters for $\pi$ and $\theta$: \\
\hspace{1cm} $\{\pi_m^{(0)},\theta_m^{(0)}\}_{m=1,\dots,M}$\\ 
\hspace{1cm} $\epsilon$: convergence threshold\\}
\KwOut{Final estimates of mixture parameters: $\widehat{\bm{\Phi}}=(\widehat{\pi}_m,\widehat{\rho}_m,\widehat{\theta}_m,\widehat{\sigma}_m)_{\ m=1,\dots,M}$}
\textbf{begin}\\
\smallskip
Compute the initial values $\{\rho_m^{(0)},\sigma_m^{(0)}\}_{m=1,\dots,M}$, for $\rho$ and $\sigma$, as\\
\setlength{\abovedisplayskip}{3pt}
\setlength{\belowdisplayskip}{3pt}
\small
\begin{equation}
\rho_m^{(0)} = \sum\limits_{x=1}^{W}\sum\limits_{y=1}^{H}h_{xy}(x\cos\theta_m^{(0)}+y\sin\theta_m^{(0)})
\label{init_rho}
\end{equation}
\begin{equation}
\sigma_m^{(0)}=\sqrt{\sum\limits_{x=1}^{W}\sum\limits_{y=1}^{H}h_{xy}\left(x\cos\theta_m^{(0)}\!+\!y\sin\theta_m^{(0)}\!-\!\rho_m^{(0)}\right)^2}
\label{init_sig}
\end{equation}
\normalsize
$t\leftarrow 0$\\
\Repeat {$\|Q(\bm{\Phi}|\bm{\Phi}^{(t)})-Q(\bm{\Phi}|\bm{\Phi}^{(t-1)})\|<\epsilon$}
{
\textbf{E-step}: Compute posterior probabilities $z_{xym}^{(t)}$ using (\ref{Estep})\\
\textbf{M-step}: Estimate mixture model parameters using (\ref{Q_it_eq})\\ $\bm{\Phi}^{(t+1)}=\arg\max\limits_{\Phi}Q(\bm{\Phi}|\bm{\Phi}^{(t)})$\\
Compute $\pi_m^{(t)}$ using (\ref{pi_est})\\
Compute $\rho_m^{(t)}$ using (\ref{rho_est})\\
Compute $u_m^{(t)}(q)_{q=1,\dots, 4}$ using (\ref{theta_opt4})\\
Select only real solutions for $u$ to obtain $u_m^{(t)}(r)_{r=1,\dots,r_{max}|r_{max}\leq 4}$\\
\For{each r}{
Compute $\theta_m^{(t)}(r)=\arctan u_m^{(t)}(r)$\;
Compute $\sigma_m^{(t)}(r)$ using $\theta_m^{(t)}(r)$ in (\ref{sig_est})\;
Compute $Q(\pi_m^{(t)},\rho_m^{(t)},\theta_m^{(t)}(r),\sigma_m^{(t)}(r))$\;
}
Deduce $r^*$ maximizing $Q$-function;\\
i.e. $r^*=\arg\max\limits_rQ(\pi_m^{(t)},\rho_m^{(t)},\theta_m^{(t)}(r),\sigma_m^{(t)}(r))$\\
Deduce the best $\sigma$ estimate: $\sigma_m^{(t)}(r^*)$\\ 
Deduce the best $\theta$ estimate: $\theta_m^{(t)}(r^*)$\\
$t\leftarrow t+1$\\
}
\smallskip
\textbf{end}
\bigskip
\end{algorithm}

\section{Experiments}
\begin{figure}[tbh!]
\centering
\subfloat[]{\includegraphics[scale=0.35]{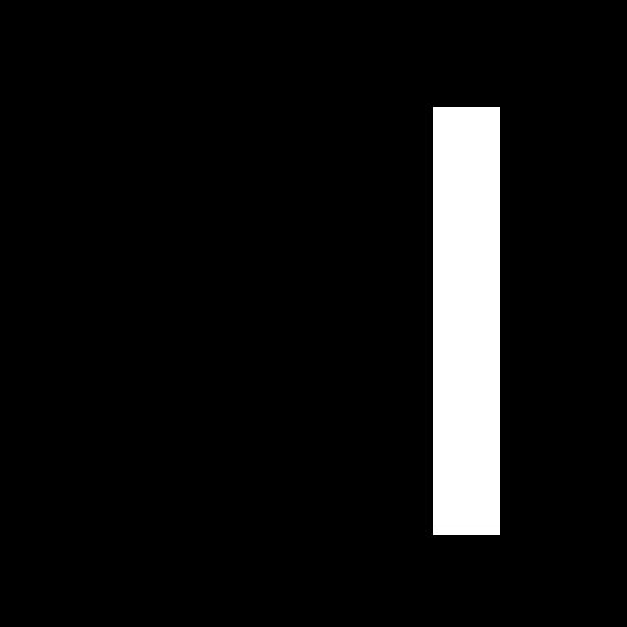}}\hspace*{0.45em}
\subfloat[]{\includegraphics[scale=0.35]{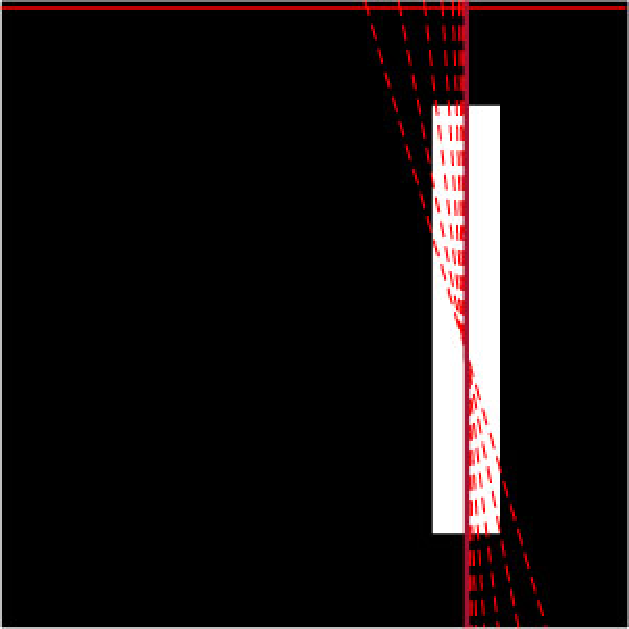}}\\
\subfloat[]{\includegraphics[scale=0.35]{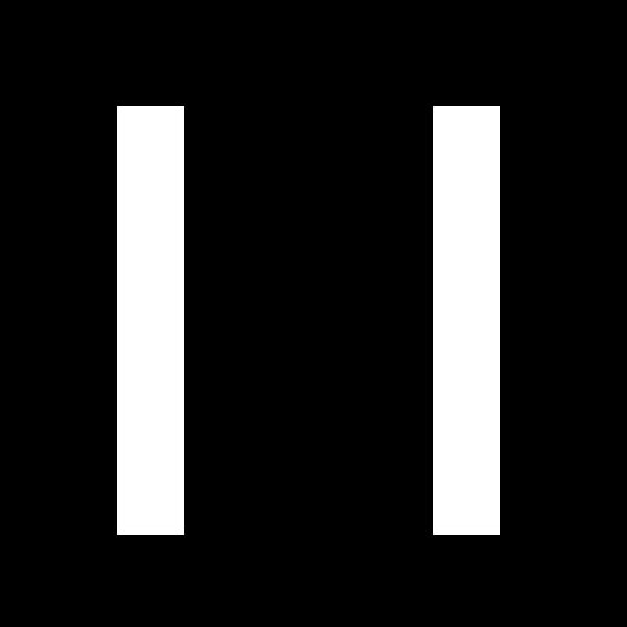}}\hspace*{0.45em}
\subfloat[]{\includegraphics[scale=0.35]{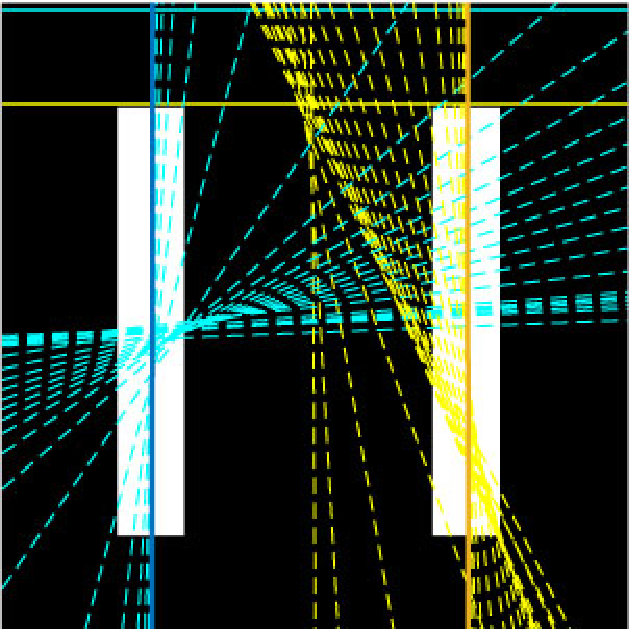}}\\
\caption{(a) and (c) Original input images representing one and two bar(s), respectively. (b) and (d) Illustration of initial, intermediate and final results, represented, respectively, by solid, dashed and bold lines in case of Algorithm 1 where, the initial angle(s) is(are) taken perpendicular to the object(s) centerline(s).}
\label{synth_ref_fig_1_2_bar}
\end{figure}

In order to show the effectiveness of the proposed linear structure detection method, experiments are carried out on a set of test images. 
The first two test images, illustrated in Figs.~\ref{synth_ref_fig_1_2_bar}a and ~\ref{synth_ref_fig_1_2_bar}c, noted, respectively, $r_1$ and $r_2$. They consist in synthetic images with size $401 \times 401$ containing, respectively, one and two thick line(s) with known values of the geometric parameters such as location, orientation and thickness: $\theta^{r_1} = 0$, $\rho^{r_1}=299$, $w^{r_1}=43$,  $\sigma^{r_1}=12.41$; $\bm{\pi}^{r_2} = [0.5 \ 0.5]$, $\bm{\theta}^{r_2} = [0 \ 0 ]$, $\bm{\rho}^{r_2}=[97 \ 299]$, $\bm{w}^{r_2}=[43 \ 43]$, $\bm{\sigma}^{r_2}=[12.41 \ 12.41]$. It is worth to note that the standard deviations corresponding to the reference image structures $\sigma^r$ are computed from $w^r$ using Eq.~\ref{final_sig_wid_rel}. The abovementioned parameters are considered as the reference parameters with which the estimated parameters, issued from Algorithms 1, are compared.

In order the show the robustness of Algorithm 1 to initialization, let us take the most unfavorable initial values for the angle and radius parameters. Then, the initial angles, taken equal to $\theta^{(0)} =\pi/2$, are perpendicular to the real orientation angle of the linear structures and the initial radius set to $\rho^{(0)} =5$ and $\rho^{(0)} =65$ are far from the real radius, given above. For Fig.~\ref{synth_ref_fig_1_2_bar}a, since it consists in a unique linear structure, the proportion parameter is then equal to $\pi^{(0)}=1$ whereas, for both figures \ref{synth_ref_fig_1_2_bar}a and ~\ref{synth_ref_fig_1_2_bar}c, the initial scale parameter $\sigma^{(0)}$ is computed through the formula Eq.~\ref{init_sig}. 
The estimated parameters vector by Algorithm 1 are noted as $\widehat{\bm{\Phi}}=(\widehat{\pi}_m,\widehat{\theta}_m,\widehat{\rho}_m,\widehat{\sigma}_m,\widehat{w}_m)_{\ m=1,\dots,M}$, where, for each linear structure $(L_m)$, the equation of the estimated centerline is given as $y^{(L_m)} = (\widehat{\rho}_m-x\cos\widehat{\theta}_m)/\sin\widehat{\theta}_m$ with $\{0\leq x\leq W(r_1),\; 0\leq y^{(L_m)}\leq H(r_1)\}$ and the width estimate $\hat{w}_m$ is directly derived from $\hat{\sigma}_m$ using Eq.~\ref{final_sig_wid_rel}. Here, the number of the mixture components is equal to 1 ($M=1$), for Fig.~\ref{synth_ref_fig_1_2_bar}a, and equals 2 ($M=2$), for Fig.~\ref{synth_ref_fig_1_2_bar}c. The convergence threshold, for the EM algorithm, is set to $\epsilon=10^{-6}$ and all the implementations are performed in Matlab environment on Dell Workstation with 2.9 GHz CPU and 32 GB RAM.

In order to compare, quantitatively, the mixture parameters estimates provided by Algorithms 1 and the reference parameters $\bm{\Phi}^{r}$, given above, two metrics are used: (1) Absolute Errors (AE) $\Delta\bm{\Phi}_i= |\bm{\Phi}_i^{r}-\widehat{\bm{\Phi}}_i|$ with $i=1,\dots,5$ and (2) Root Mean Square Error (RMSE) computed as \small{$\mathop{\mathlarger{\sum}}_{\Delta\bm{\Phi}_i}=\sqrt{\mathop{\mathlarger{\sum}}_{m=1}^M\Delta\Phi_{i,m}^2/M}$} \normalsize where, the symbol $\ \widehat{}\ $ is used to indicate the parameters estimated values.  

The AEs between the real and the estimated geometric parameters for Figs.~\ref{synth_ref_fig_1_2_bar}a and ~\ref{synth_ref_fig_1_2_bar}c are given as follows: Fig.~\ref{synth_ref_fig_1_2_bar}a: $\Delta{\theta}=3\times 10^{-4}$, $\Delta{\rho}=0.07$, $\Delta{\sigma}=3\times 10^{-3}$ and $\Delta{w}=0.01$; Fig.~\ref{synth_ref_fig_1_2_bar}c: $\Delta\bm{\pi}=[0 \ 0]$, $\Sigma_{\Delta\bm{\pi}}=0$, $\Delta\bm{\theta}=[0.004 \ 0.002]$, $\Sigma_{\Delta\bm{\theta}}=2.9\times 10^{-4}$, $\Delta\bm{\rho}=[0.08 \ 0.05]$, $\Sigma_{\Delta\bm{\rho}}=0.07$, $\Delta\bm{\sigma}= [0.003 \ 0.003]$, $\Sigma_{\Delta\bm{\sigma}}=3\times 10^{-3}$, $\Delta\bm{w}=[0.01 \ 0.01]$, $\Sigma_{\Delta\bm{w}}=0.01$.

Initial, intermediate and final centerlines evolved by Algorithm 1, for both input images, are illustrated in Figs.~\ref{synth_ref_fig_1_2_bar}b and \ref{synth_ref_fig_1_2_bar}d. 

It appears from the above-mentioned estimation errors and figures that the object centerlines and thicknesses are accurately recovered where, the estimation errors are almost zero and this, even if the radius and orientation angle parameters are initialized to values as far as possible from the true values.  

\begin{figure}[tbh!]
\centering
\includegraphics[scale=0.7]{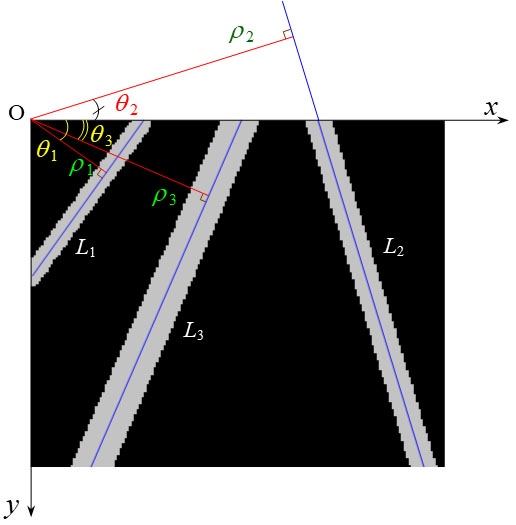}
\caption{Radius and orientation angle parameters for an input synthetic image representing 3 thick lines.}
\label{synth_ref_fig}
\end{figure}

The third test image, illustrated in Fig.~\ref{synth_ref_fig}, consists in a synthetic image with size $169 \times 142$, noted $r_3$, which contains three linear structures ($M=3$) with known values of their geometric parameters such as location, orientation and thickness: $\bm{\pi}^{r_3} = [0.14 \ 0.34 \ 0.52]$; $\bm{\theta}^{r_3} = [35 \ -17 \ 23]$; $\bm{\rho}^{r_3}=[38 \ 112 \ 79]$; $\bm{w}^{r_3}=[8 \ 10 \ 15]$ and $\bm{\sigma}^{r_3}=[2.3 \ 2.9 \ 4.3]$. As for the above used test images, the values of $\sigma^{r_3}$ are computed from $w^{r_3}$ using Eq.~\ref{final_sig_wid_rel}. 

In this experiment, with the aim to test the robustness of the proposed algorithms against blur, noise and the combination of both, in addition to the utilization of the input noise-free image, corrupted versions of the latter by blur and various levels of additive white noise are also used. While the blurred image is generated by convolving the input sharp image in Fig.~\ref{synth_ref_fig} with a Gaussian kernel $\Theta(x,y;\kappa)=(2\pi\kappa^2)^{-1/2}\exp(-(x^2+y^2)/(2\kappa^2))$ of size $15\times 15$ with spread $\kappa (\kappa=3)$, the noised image is achieved by adding a white Gaussian noise (AWGN) $\epsilon\rightarrow\mathcal{N}(0,\sigma_n^2)$ with increasing standard deviation $(\sigma_n = 50, 100, 150)$ to the original or the blurred images. 

About the EM algorithm initialization, the mixture parameters are set as follows: (1) the components of the mixture are taken in equal proportions, i.e. ${\pi_m}^{(0)}=\frac{1}{M},\ m=1,\dots,M$. (2) The orientation angle of the first  component $(m=1)$ is initialized randomly, i.e. $\theta_1^{(0)} =\pi r_d$ with $r_d$ is a random real value in $[0 \ 1]$ and the initial angles for the remainder of the mixture components are divided equally on $[\theta_1^{(0)} \; \theta_1^{(0)}\!+\!\pi]$, providing: $\bm{\theta}^{(0)}=[\theta_1^{(0)} \quad \theta_1^{(0)}\!+\!\frac{\pi}{M} \quad \theta_1^{(0)}\!+\!\frac{2\pi}{M} \quad \dots \quad \theta_1^{(0)}\!+\!\frac{(M-1)\pi}{M}]$. In our case, $M=3$ and $\bm{\theta}^{(0)}=[\theta_1^{(0)} \quad \theta_1^{(0)}\!+\!\frac{\pi}{3} \quad \theta_1^{(0)}\!+\!\frac{2\pi}{3}]$. (3) The initial radius and standard deviation parameters $\rho_m^{(0)}$ and $\sigma_m^{(0)}, \ m=1,\dots,M$, are obtained from Eq.~\ref{init_rho} and Eq.~\ref{init_sig}, respectively. 

\begin{figure}
\centering
\subfloat[$\sigma_n=0,\kappa=0$]{\includegraphics[scale=0.58]{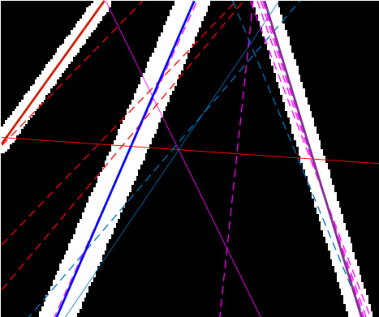}}\hspace*{0.2em}
\subfloat[$\sigma_n=50,\kappa=3$]{\includegraphics[scale=0.58]{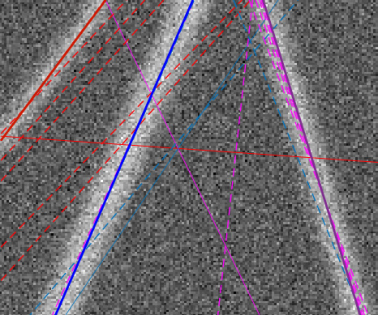}}\\
    \subfloat[$\sigma_n=100,\kappa=3$]{\includegraphics[scale=0.58]{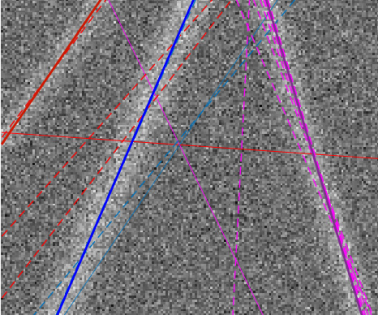}}\hspace*{0.2em}
\subfloat[$\sigma_n=150,\kappa=3$]{\includegraphics[scale=0.58]{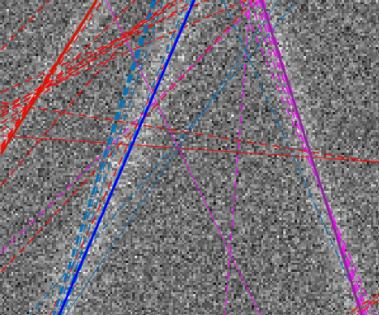}}\\
     \caption{Illustration of initial, intermediate and final results, represented, respectively, by solid, dashed and bold lines in case of Algorithm 1 using angle random initialization.}
    \label{syn_rand_init_alg1}
\end{figure}

Relative Absolute Errors (RelAE) for the thickness parameter is also computed where, $\Delta^{rel}\bm{w}= \frac{\Delta\bm{w}}{\bm{w}^{r_3}}$. The line detection accuracy, the computational time and the number of iterations are reported in Table~\ref{test3_resul_alg1}. The linear structure centerlines detected by Algorithm 1 applied on the noise-free input image and its noisy/blurry versions are illustrated in Fig.~\ref{syn_rand_init_alg1}.  

\begin{table*}[tbh!]
\small
\centering
\caption{Error values between real and estimated parameters using Algo 1 with random initial $\theta$ for Fig.~\ref{synth_ref_fig} in function of blur and noise. The component ranks are given in Fig.~\ref{synth_ref_fig}.}
\label{test3_resul_alg1}
\begin{tabular}{c|c|c|c|c}
$(\sigma_n,\kappa)$&(0,0) & (50,3) & (100,3) & (150,3)\\ 
\hline
\# iter & 66 & 75 & 96 & 120 \\
runtime & 0.26 s & 0.30 s & 0.38 s & 0.48 s \\ 
$\Delta\bm{\pi}$ & 0 \quad 0 \quad 0 & 0 \ \quad 0 \ \quad 0 & 0.00 \ 0.00 \ 0.00 & 0.00 \ 0.00 \ 0.00\\ 
$\Sigma_{\Delta\bm{\pi}}$  & 0 & 0  & 0.00  & 0.00 \\ 
$\Delta\bm{\theta} (\degree)$ & 0 \quad 0 \quad 0 & 1.10 \ 0.14 \ 0.09 & 1.30 \ 0.18 \ 0.08 & 2.38 \ 0.67 \ 0.36\\ 
$\Sigma_{\Delta\bm{\theta}} (\degree)$ & 0 & 0.64 & 0.76 & 1.44 \\ 
$\Delta\bm{\rho}$ & 0  \quad 0 \quad 0 & 0.41 \ 0.09 \ 0.01 & 0.53  \  0.70  \;  0.06 & 1.32 \ 1.30 \ 0.74 \\ 
$\Sigma_{\Delta\bm{\rho}}$ & 0 & 0.24 & 0.51 & 1.15\\ 
$\Delta\bm{\sigma}$  &0 \quad 0 \quad 0 & 0.79  \  0.36  \  0.84 & 1.78  \  1.91  \  0.19  & 1.63 \ 1.58 \ 1.62\\ 
$\Sigma_{\Delta\bm{\sigma}}$  & 0 &0.7 & 1.15 & 1.61  \\ 
$\Delta\bm{w}$  & 0 \quad 0 \quad 0 & 2.76  \  1.24  \  2.92  & 6.15 \   6.62  \  0.67 & 5.65 \ 5.48 \ 5.64\\ 
$\Delta^{rel}\bm{w}$  &  0 \quad 0 \quad 0 & 0.34  \ 0.19 \ 0.19 & 0.77 \ 0.64 \ 0.04 & 0.71 \ 0.53 \ 0.37 \\ 
$\Sigma_{\Delta\bm{w}}$ & 0 & 2.43  & 5.23 & 5.59 \\
\end{tabular}
\end{table*}

In light of the obtained results, the following remarks can be made:

(1) When the blur effect and white additive noise are added to noise/blur-free input image, the number of iterations and runtime increase and are, then, proportional to the intensity of the added noise. This observation may be intuitive since the blur and noise affect the data. 

\begin{figure}[tbh!]
\centering
\subfloat[]{\includegraphics[scale=0.18]{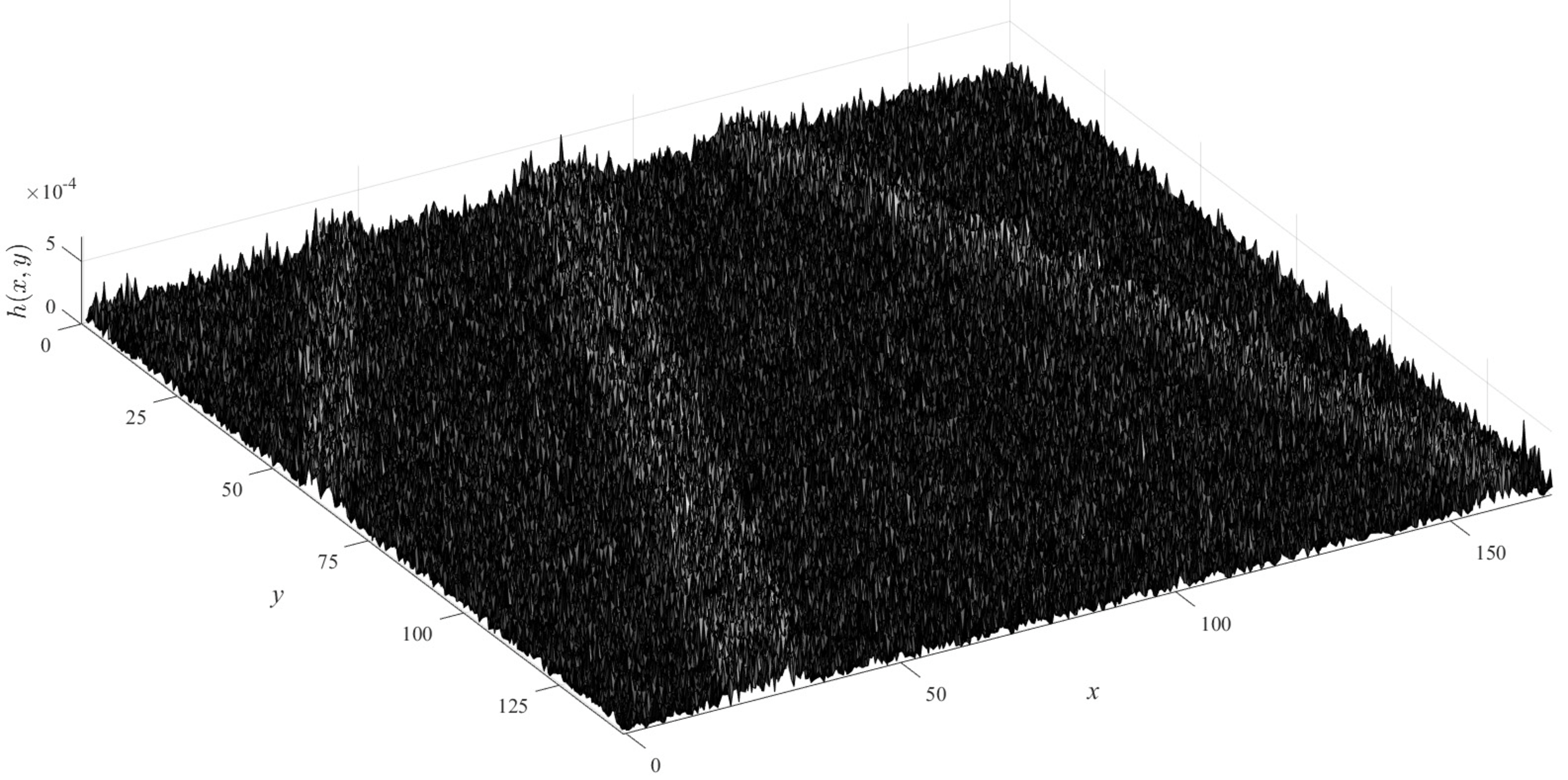}}\hspace*{0.4em}\\
\subfloat[]{\includegraphics[scale=0.18]{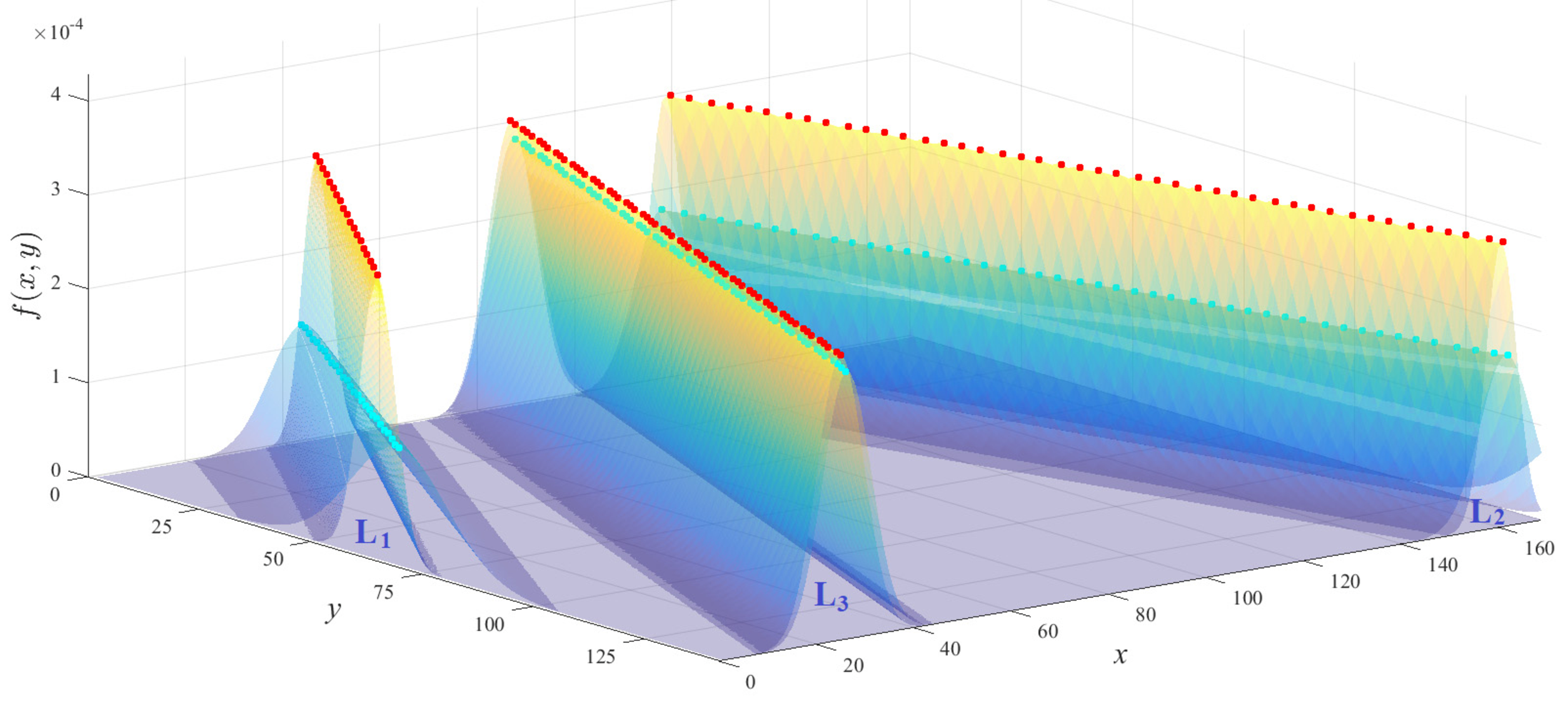}}\hspace*{0.4em}\\
\subfloat[]{\includegraphics[scale=0.18]{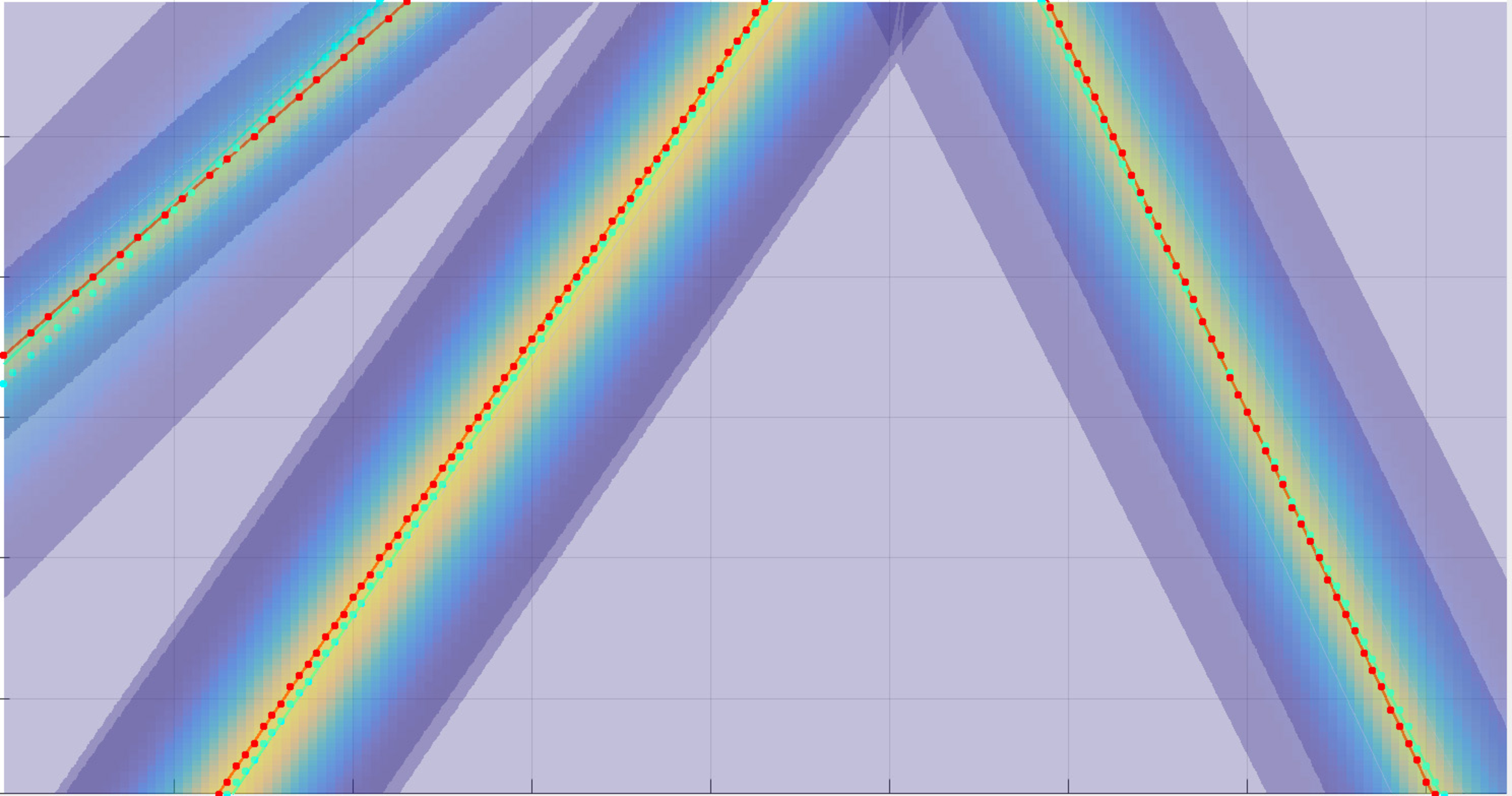}}
\caption{(a) 3D view of the noisy image. (b) Real linear anchored Gaussians (maxima in red) versus computed linear anchored Gaussians (maxima in cyan) for input image in (a). (c) Top view of figure in (b) highlighting the deviations in centerlines positions where, the computed centerline for the structure $L_1$ presents the highest position error as mentioned in discussion.}
\label{mix_fitt}
\end{figure}

(2) The geometric parameters of the noise/blur-free objects contained in the input image are accurately estimated where, the errors are null, as shown in the first column of Table~\ref{test3_resul_alg1}. By adding the blur effect to the original input image and increasing the noise intensity, small but not negligible errors appear for the detection of the linear structure $L_1$ (see Fig.~\ref{synth_ref_fig}) and, to a lesser degree, for the structure $L_2$. In fact, in presence of blur and strong noise, the biggest estimation errors occur on the thickness parameter $w$ of which the maximal RelAEs, for the object $L_1$, $L_2$ and $L_3$ reach 77\%, 64\% and 37\% respectively. Obviously, these errors concern also the scale parameter $\sigma$ since these parameters are linked by Eq.~\ref{final_sig_wid_rel}. These large error values can be explained by the fact that the mixture model using EM algorithm attempts to use all the available data, including the noisy background. In other words, to gather the data, the 1D profiles of the linear anchored Gaussians (see Fig.~\ref{l_Gauss}) get wider, which increase the scale parameter values $\sigma$ for all the linear structures present in the image. Thus, for the case $(\sigma_n = 150,\kappa=3)$, the $\sigma$ parameter values reach $\bm{\sigma}=[3.93\; 4.59\; 6.07]$ exceeding the actual $\sigma$ by a difference vector equal to $\Delta\bm{\sigma}=[1.63 \; 1.58 \; 1.62]$ of which the $1^{st}$, $2^{nd}$ and $3^{rd}$ components correspond to the linear objects $L_1$, $L_2$ and $L_3$, respectively.  
In the other hand, the maximal AEs in orientation $\theta$ and radius $\rho$, for the object $L_1$, reach $2.38\degree$ and 1.32 pixels, respectively; whereas, they are, respectively, about $0.67\degree$ and 1.3 pixels, for the object $L_2$. However, the linear structure $L_3$, presents the best results where, in presence of blur and noise, the AEs approximate $0.36\degree$ and 0.74 pixels, for $\theta$ and $\rho$, respectively. The best accuracy on the centerline localization of the structure $L_3$ is, among other, due to its position in the image. Indeed, even if $\sigma$ value is important, the object L3 is almost in the middle of the image and is relatively far from the borders which makes it less sensitive to border effects. So, the real linear anchored Gaussian fitting the structure $L_3$ is totally held by the image support $\mathcal{D}$, contrary to the real Gaussian distributions fitting the structures $L_2$ and $L_3$ where, some parts of them are outside the domain $\mathcal{D}$, as illustrated in Fig.~\ref{mix_fitt}. During computation, because of the noisy background, the Algorithm 1 attempts to bring the estimated Gaussians into the image domain so that it maximizes the mixture likelihood function. Thus, for structures $L_1$ and $L_2$, which are near the boundaries, their computed centerlines present some deviation from the real computed linear anchored Gaussian maxima, as read in Table~\ref{test3_resul_alg1} and shown in Fig.~\ref{mix_fitt}. We will present in the next subsection a method to address the problem dealing with a noisy or a complex background. 

\subsection{Dynamic background subtraction}
Usually, in object detection applications, the image background can present irrelevant parts (e.g. artifacts, noise, etc.) being able to affect negatively the linear structure fitting via the proposed Algorithm 1. Then, dismissing the irrelevant data when running the EM algorithm will unavoidably contribute in a best linear structure fitting. That is why, we propose in this subsection a modified version of Algorithm 1, called Algorithm 2. 
In the latter, the background is subtracted from the input image so that only the image bands of which the number is equal to the number of the mixture components, are considered. Each image band is delimited by the two lines parallel to the computed centerline and distanced from it by $\pm\nu\sigma$. 

Since the width $w$ of the linear structure is linked to $\sigma$ through Eq.\ref{final_sig_wid_rel}, the value of $\nu$ should be greater than $\sqrt{3}$ to guarantee that all the structure pixels are included in the image band. Then, a value of 2 for $\nu$ could be sufficient. 

\begin{algorithm}
\KwIn{$h$: Normalized input grayscale image\\
\hspace{1cm} $M$: Number of mixture components\\
\hspace{1cm} Initial mixture parameters for $\pi$ and $\theta$: \\
\hspace{1cm} $\{\pi_m^{(0)},\theta_m^{(0)}\}_{\ m=1,\dots,M}$\\ 
\hspace{1cm} $\epsilon$: convergence threshold\\}
\KwOut{Final estimates of mixture parameters: $\widehat{\bm{\Phi}}=(\widehat{\pi}_m,\widehat{\rho}_m,\widehat{\theta}_m,\widehat{\sigma}_m)_{\ m=1,\dots,M}$}
\textbf{begin}\\
\smallskip
\dots\\
Same computations as in Algorithm 1.\\
\dots\\
$t\leftarrow 0$\\
\Repeat {$\|Q(\bm{\Phi}|\bm{\Phi}^{(t)})-Q(\bm{\Phi}|\bm{\Phi}^{(t-1)})\|<\epsilon$}
{
\For{each m}{
Compute the half-top region: $K_m^t|K_t(x,y)=(x\cos\theta_m+y\sin\theta_m\geq \rho_m-2\sigma_m)$;
Compute the half-bottom region: $K_m^b|K_b(x,y)=(x\cos\theta_m+y\sin\theta_m\leq \rho_m+2\sigma_m)$;
Compute the region embedding the $m^{th}$ linear structure: $K_m = K_m^t\cap K_m^b$;  
}
Deduce the input image after background subtraction: $I_{bs} = \bigcup_{m=1}^M K_m$;\\
Deduce the new normalized input image: $h_{bs}(x,y)=I_{bs}(x,y)\Big/\sum_{x=1}^{W}\sum_{y=1}^{H}I_{bs}(x,y)$\\
E-step: as in Algorithm 1, compute the posterior probabilities $z_{xym}^{(t)}$ using (\ref{Estep})\\
M-step: as in Algorithm 1, estimate the mixture model parameters using (\ref{Q_it_eq})\\ $\bm{\Phi}^{(t+1)}=\arg\max\limits_{\Phi}Q(\bm{\Phi}|\bm{\Phi}^{(t)})$\\
\dots\\
\dots\\
Repeat the steps used in Algorithm 1.\\
\dots\\
\dots\\
$t\leftarrow t+1$
}
\smallskip
\textbf{end}
\bigskip
\label{}
\caption{EM algorithm for linear anchored Gaussian mixture model with dynamic background subtraction}
\end{algorithm}

\begin{figure}[h]
\centering
\subfloat[$\sigma_n=0,\kappa=0$]{\includegraphics[scale=0.2]{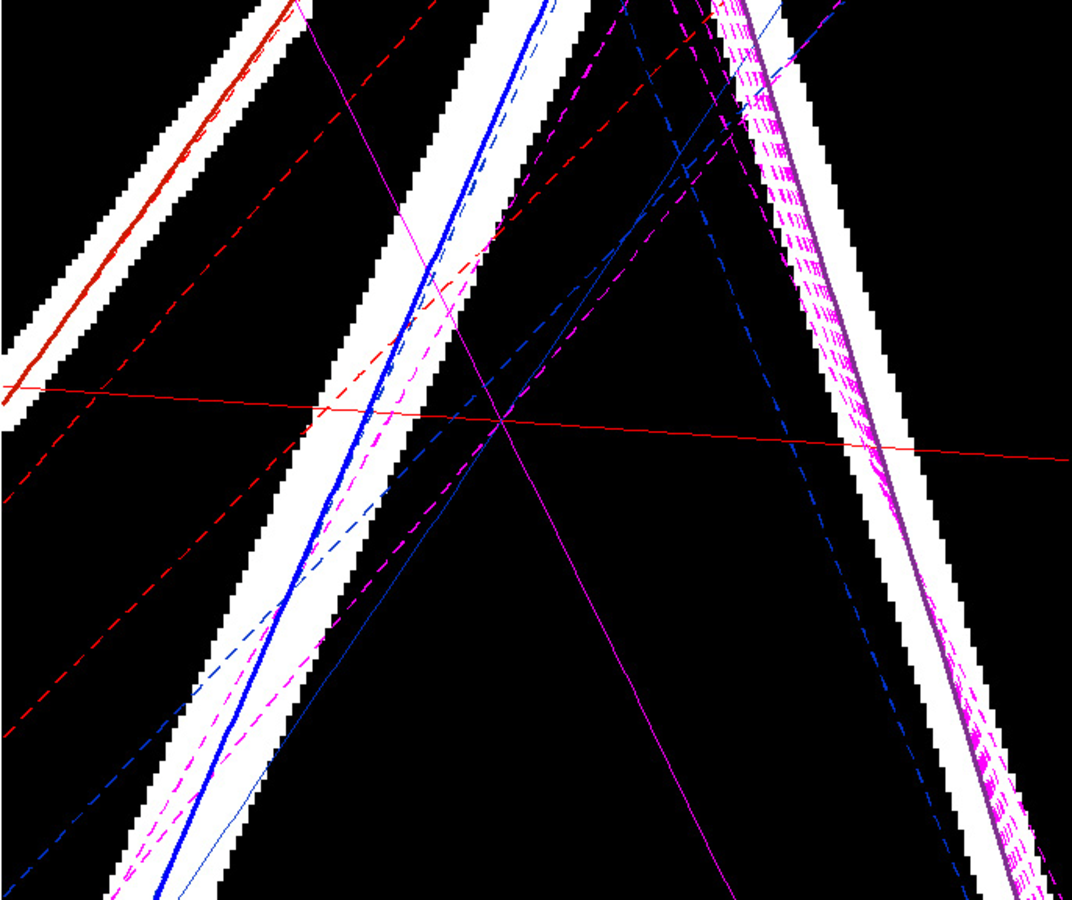}}\hspace*{0.2em}
\subfloat[$\sigma_n=50,\kappa=3$]{\includegraphics[scale=0.2]{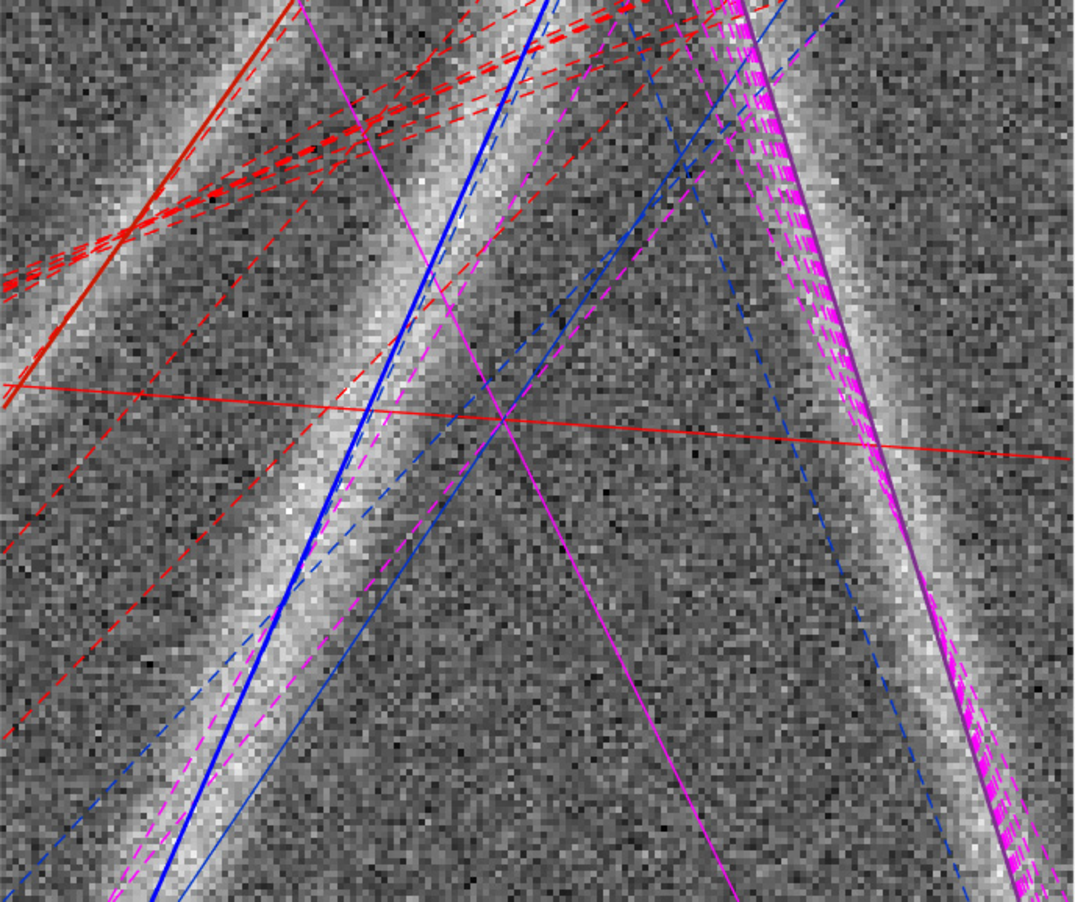}}\\
\subfloat[$\sigma_n=100,\kappa=3$]{\includegraphics[scale=0.2]{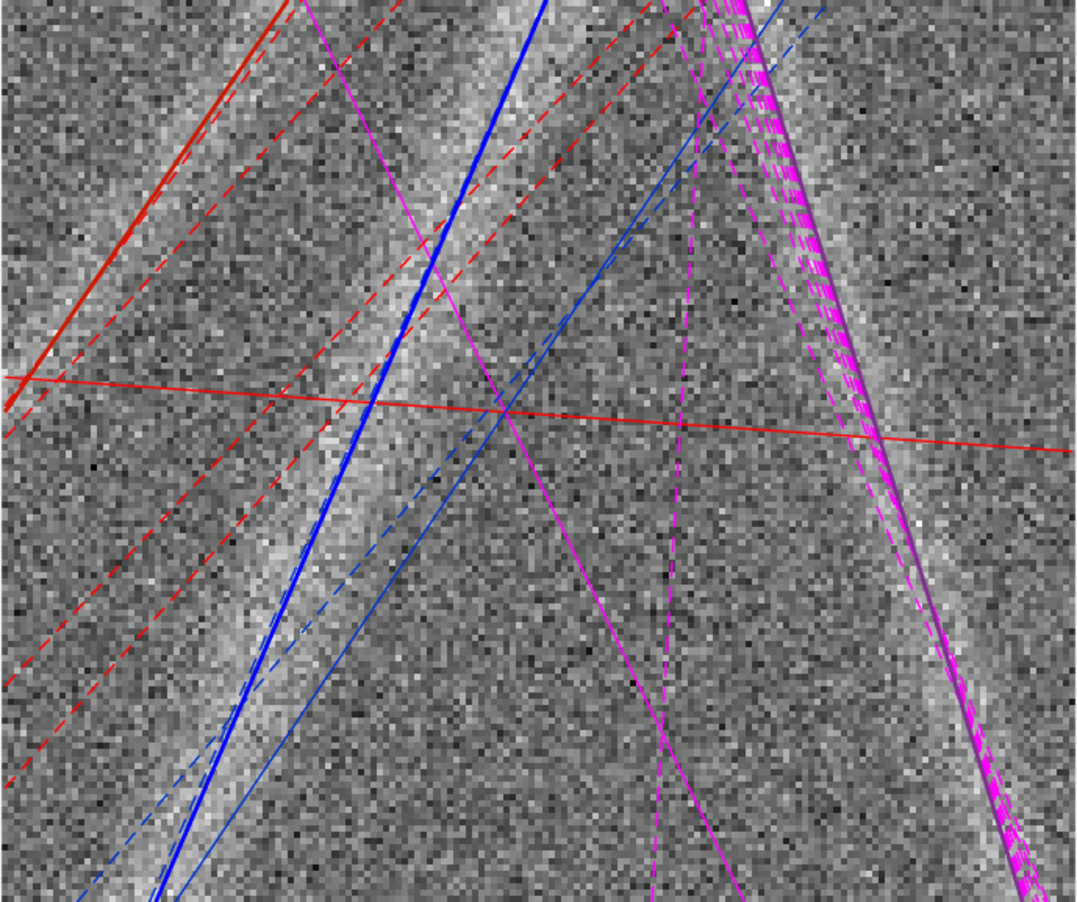}}\hspace*{0.2em}
\subfloat[$\sigma_n=150,\kappa=3$]{\includegraphics[scale=0.2]{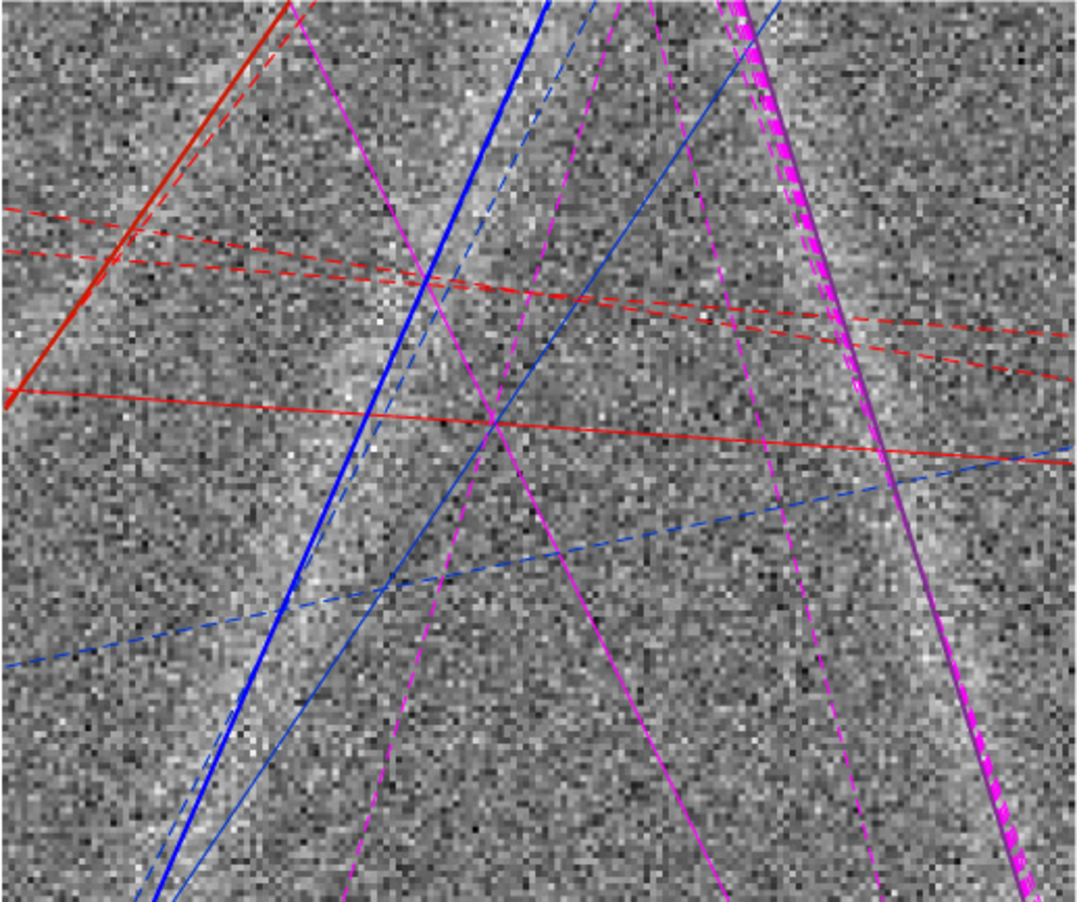}}\\
\caption{Illustration of initial, intermediate and final results, represented, respectively, by solid, dashed and bold lines in case of Algorithm 2 using angle random initialization.}
\label{syn_rand_init_alg2}
\end{figure}

The line detection accuracy, the computational time and the number of iterations are reported in Table~\ref{test3_resul_alg2}. The linear structure centerlines detected by Algorithm 2 applied on the noise-free input image and its noisy/blurry versions are illustrated in Fig.~\ref{syn_rand_init_alg2}. 

\begin{table*}[tbh!]
\small
\centering
\caption{Error values between real and estimated parameters using Algo 2 with random inial $\theta$ for Fig.~\ref{synth_ref_fig} in function of blur and noise. The component ranks are given in Fig.~\ref{synth_ref_fig}.}
\label{test3_resul_alg2}
\begin{tabular}{c|c|c|c|c}
$(\sigma_n,\kappa)$&(0,0) & (50,3) & (100,3) & (150,3)\\ 
\hline
\# iter & 78 & 86 & 106 & 119 \\
runtime & 0.36 s & 0.38 s & 0.46 s & 0.53 s \\ 
$\Delta\bm{\pi}$ & 0 \quad 0 \quad 0 & 0 \ \quad 0 \ \quad 0 & 0.00 \ 0.00 \ 0.00 & 0.00 \ 0.00 \ 0.00\\ 
$\Sigma_{\Delta\bm{\pi}}$  & 0 & 0  & 0.00  & 0.00 \\ 
$\Delta\bm{\theta} (\degree)$ & 0 \quad 0 \quad 0 & 0.15 \ 0.18 \ 0.25 & 1.06 \ 0.17 \ 0.02 & 1.37 \ 0.22 \ 0.05  \\        
$\Sigma_{\Delta\bm{\theta}} (\degree)$ & 0 & 0.18 & 0.62 & 0.80 \\ 
$\Delta\bm{\rho}$ & 0  \quad 0 \quad 0 & 0.31 \ 0.28 \ 0.20 & 0.19 \ 0.22 \ 0.02  & 0.86 \ 0.26 \ 0.19 \\                     
$\Sigma_{\Delta\bm{\rho}}$ & 0 & 0.27 & 0.17 & 0.53\\ 
$\Delta\bm{\sigma}$  &0 \quad 0 \quad 0 & 0.35 \ 0.32 \ 0.18 & 0.50 \ 0.17 \ 0.06 & 0.49 \ 0.38 \ 0.29\\                        
$\Sigma_{\Delta\bm{\sigma}}$  & 0 & 0.29 & 0.31 & 0.39 \\ 
$\Delta\bm{w}$  & 0 \quad 0 \quad 0 & 1.27 \ 1.11 \ 0.62 & 1.73 \ 0.59 \ 0.22 & 1.70 \ 1.30 \ 0.99 \\                 
$\Delta^{rel}\bm{w}$  &  0 \quad 0 \quad 0 & 0.15 \ 0.11 \ 0.04 & 0.28 \ 0.06 \ 0.01 & 0.21 \ 0.12  \ 0.06 \\                   
$\Sigma_{\Delta\bm{w}}$ & 0 & 1.02 & 1.06 & 1.36 \\
\end{tabular}
\end{table*}

As remarked in the last table and figures, a substantial improvement on the parameters estimation of the linear structures is obtained using Algorithm 2. In fact, for example, with the presence of a strong noise and blur ($\sigma_n = 150, \kappa=3$), the errors in objects positions ($\theta,\rho$) and widths ($w$) drop from the values $\Delta\hat{\bm{\theta}}=\{2.38 \; 0.67 \; 0.36\}$, $\Delta\hat{\bm{\rho}}=\{1.32 \; 1.30 \; 0.74\}$ and $\Delta^{rel}\hat{\bm{w}}=\{0.71 \; 0.53 \; 0.37\}$ obtained by Algorithm 1 to the values $\Delta\hat{\bm{\theta}}=\{1.37 \;0.22 \; 0.05\}$, $\Delta\hat{\bm{\rho}}=\{0.86 \; 0.26 \; 0.19\}$ and $\Delta^{rel}\hat{\bm{w}}=\{ 0.21 \; 0.12  \; 0.06\}$, provided by Algorithm 2. For Algorithm 2, the structure detection improvement is expected since the irrelevant parts of background are discarded and then, only the objects of interest and their closest neighborhood are kept. That is why, for example, despite its little area, which constitutes a challenging situation in most mixture model paradigms, the structure $L_1$ parameters are, here, accurately recovered.     

Until now, all the experiments are done assuming that the number of the mixture model components $M$ are known. However, in real world applications, this parameter is often unknown. That is why, we propose in the following experimental part to apply the Hessian of the image in order to derive the components number $M$ from the computed Hessians orientations as done in \cite{Goumeidane2021}. In addition, this Hessian-based method provides the initial values of the orientation angle $\theta$.         
To sum up the method, the multiscale Hessian is applied on the image to highlight the strongest Hessian responses of the structures. The highlighted response directions are exploited to construct an histogram of orientations. The arguments of the histogram peaks are the initial orientation angles. In case of presence of parallel structures, one orientation can concern several structures. To find how much structures are related to each orientation, some orientation maps are used. For more details, please refer to \cite{Goumeidane2021}. 

\begin{figure}[tbh!]
\centering
\subfloat[]{\includegraphics[scale=0.5]{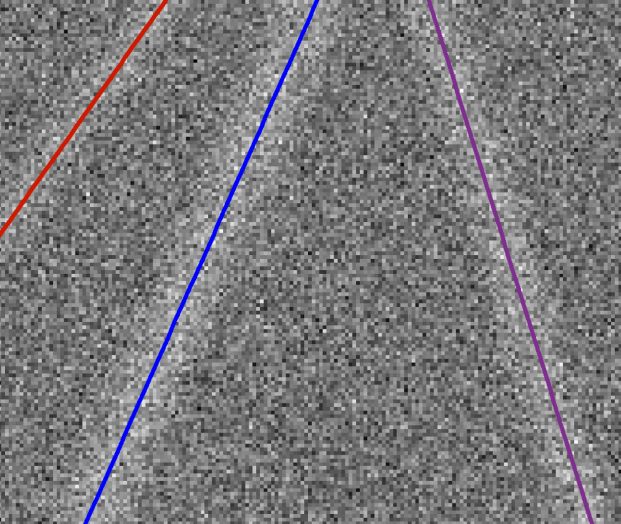}}\\
\subfloat[]{\includegraphics[scale=0.46]{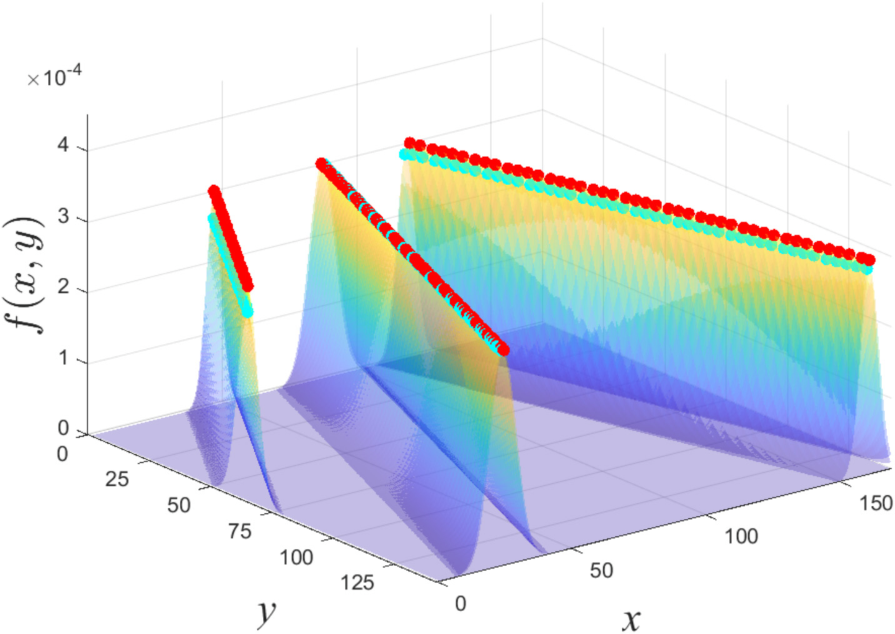}}\\
\caption{(a) Final linear structure centerline estimates, in presence of noise and blur ($\sigma_n=150,\kappa=3$), using Algorithm 2 with Hessian-based angle initialization. (b) Real linear anchored Gaussians (maxima in red) versus computed linear anchored Gaussians (maxima in cyan), where the mixture fitting errors are practically insignificant.}
\label{syn_Hess_init_alg2}
\end{figure}

In Table~\ref{test3_alg2_hess} and Fig.~\ref{syn_Hess_init_alg2}, are depicted the results of applying Algorithm 2 on the test image in Fig.~\ref{synth_ref_fig}, corrupted by additive white noise and blur effect with values $\sigma_n = 150$ and $\kappa=3$, respectively. Here, the number of the mixture model components, which is considered unknown, is computed by the Hessian-based method giving $M$ equal to 3; and, at meantime, this method computes the initial orientation angles vector as $\bm{\theta}^{(0)}_{Hess} = [37\degree \; -16\degree \; 23\degree ]$ of which the mixing components values correspond to the linear structures $L_1$, $L_2$ and $L_3$, respectively. 
We can remark that the Hessian of the image permits to make a good prediction of the orientation angles of the linear objects contained in the image and the number of these objects, despite the presence of a pronounced blur effect and a strong additive noise.

The best approximation of $\bm{\theta}^{(0)}$, using the image Hessian, contributes greatly to a rapid convergence of Algorithm 2 to the optimal mixture parameters, where only 86 iterations are required to reach the convergence threshold $\epsilon$, spending less than 0.4 seconds. By comparing the estimation errors, provided in Table~\ref{test3_alg2_hess} to those given in Tables~\ref{test3_resul_alg1} and \ref{test3_resul_alg2}, we observe clearly that, besides an efficient model selection of the finite linear anchored Gaussian mixture model, the best estimates of the latter, are obtained by the multiscale image Hessian-based paradigm, as ascertained also by Figs.~\ref{syn_Hess_init_alg2}a and \ref{syn_Hess_init_alg2}b.   

\begin{table}[tbh!]
\small
\centering
\caption{Error values between real and estimated parameters using Algorithm 2 with Hessian-based initial $\theta$ for Fig.~\ref{synth_ref_fig} with noise and blur ($\sigma_n = 150$, $\kappa=3$)}
\label{test3_alg2_hess}
\begin{tabular}{c|c}
& $\sigma_n = 150, \kappa=3$ \\ 
\hline
\# of iterations \;|\; Runtime & 86 \;|\; 0.39 s\\
\# of components ($M$) & 3 \\ 
$\Delta\bm{\pi}$\;|\;$\Sigma_{\Delta\bm{\pi}}$ & [0.00 \; 0.00 \; 0.00] \;|\; 0.00\\ 
$\Delta\bm{\theta} (\degree)$ \;|\;$\Sigma_{\Delta\bm{\theta}} (\degree)$& [0.26 \; 0.31 \; 0.43] \;|\; 0.34 \\        
$\Delta\bm{\rho}$\;|\; $\Sigma_{\Delta\bm{\rho}}$& [0.19 \; 0.54 \; 0.38] \;|\; 0.40 \\                 
$\Delta\bm{w}$\;|\; $\Sigma_{\Delta\bm{w}}$ & [0.25 \; 0.77 \; 0.17] \;|\; 0.48\\                               
$\Delta^{rel}\bm{w}$  & [0.03 \; 0.07  \; 0.01] \\  
\end{tabular}
\end{table}

\subsection{Linear structure objects detection in real-world images}
\begin{figure}[tbh!]
\centering
\subfloat[]{\includegraphics[scale=0.306]{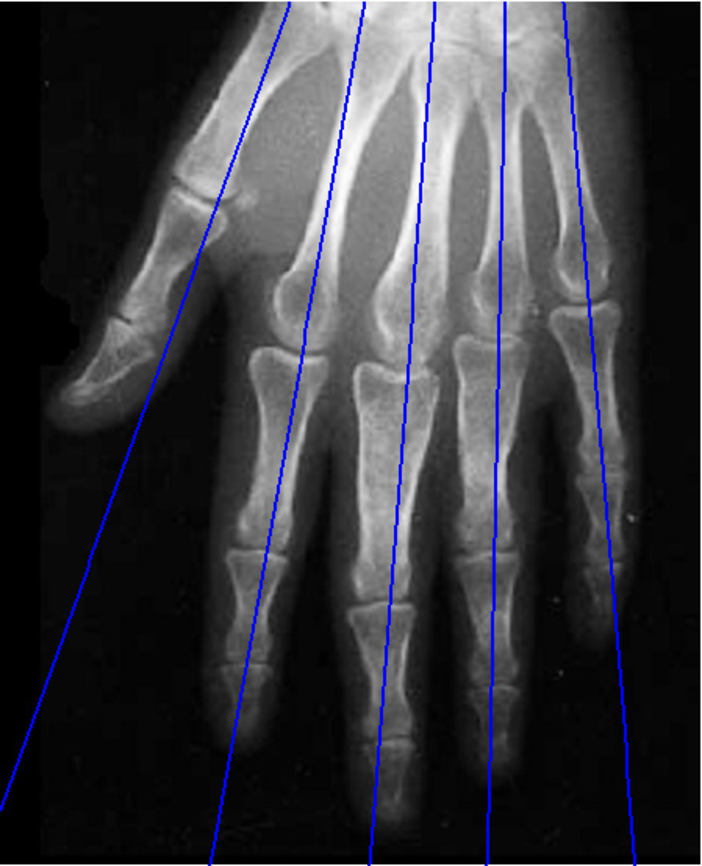}}\hspace*{0.4em}
\subfloat[]{\includegraphics[scale=0.3145]{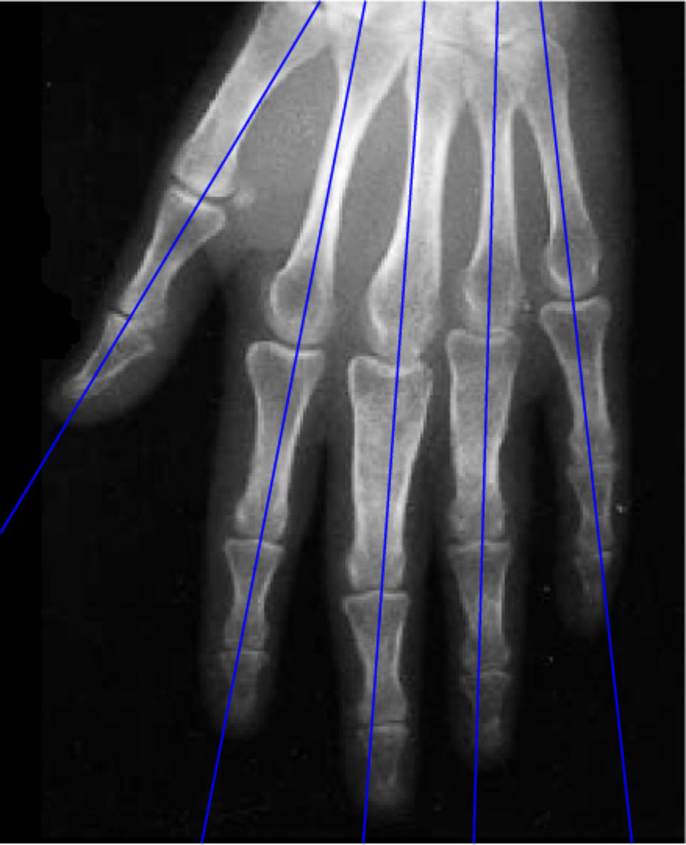}}\\
\caption{Estimated linear structures, formed by the phalanges and metacarpals bones in a hand X-ray image, using (a) Algo 1 and (b) Algo 2.}
\label{Hand_Xray}
\end{figure}

\begin{figure*}[tbh!]
\centering
\subfloat[]{\includegraphics[scale=0.32]{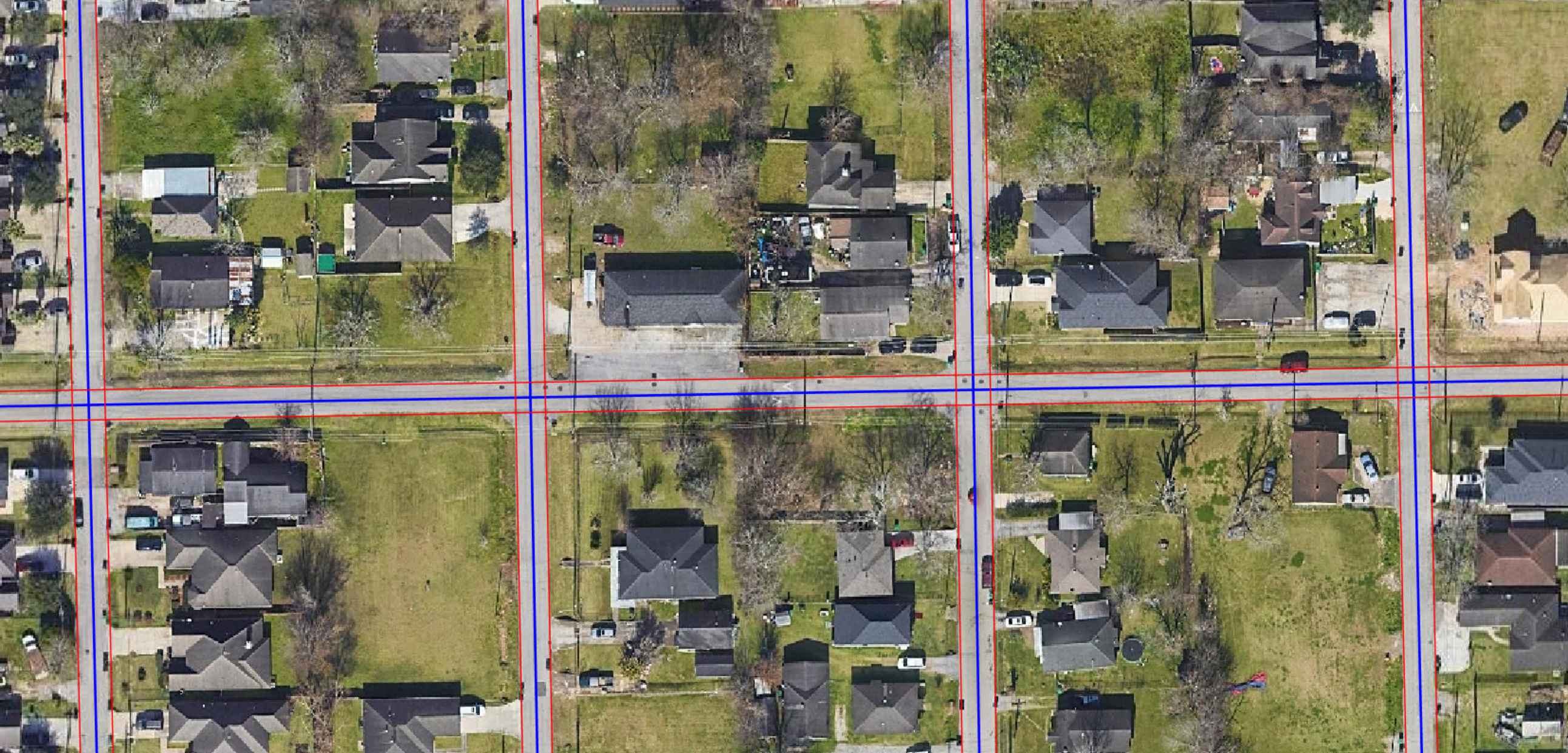}}\\ 
\subfloat[]{\includegraphics[scale=0.28]{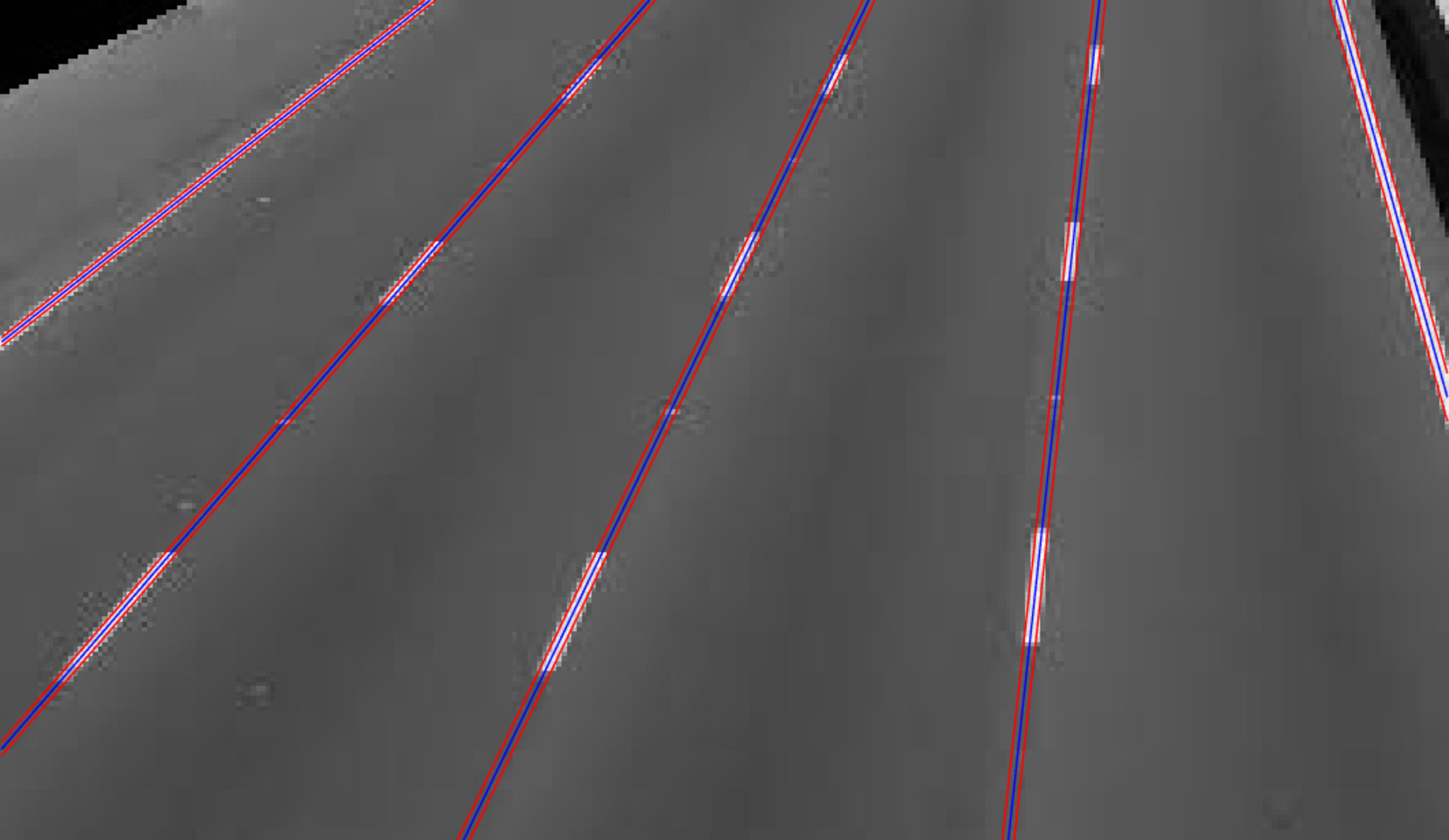}}\\
\subfloat[]{\includegraphics[scale=0.28]{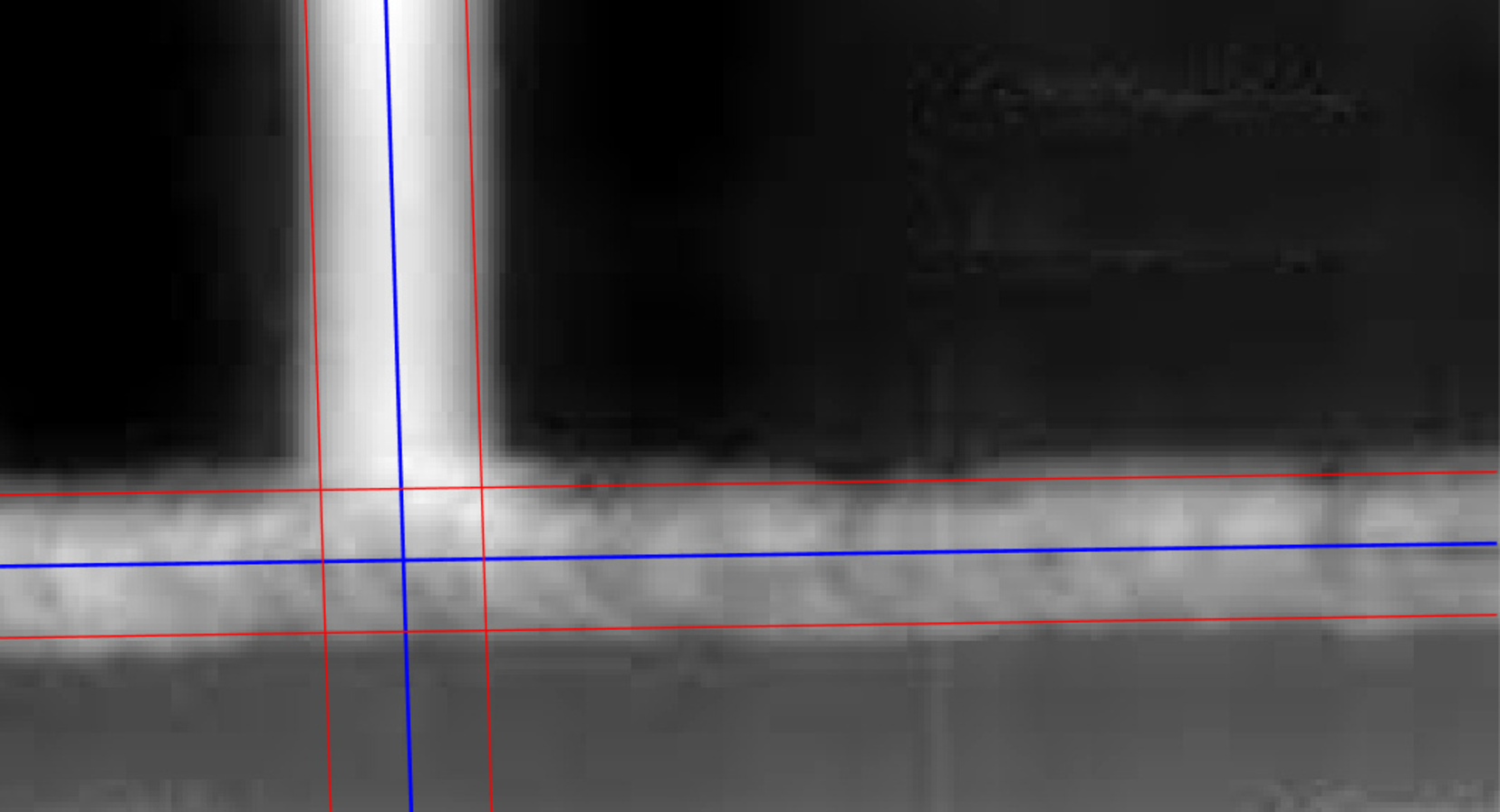}}
\caption{Results obtained by Algo 2 on real images, where the blue line highlights the structure centerline and the pair of red lines delineates the structure thickness.}
\label{real_im_res}
\end{figure*}

As mentioned in the Introduction, an accurate detection of the centerlines of linear structure objects is a challenging topic in many sensitive real-world applications such X-ray imaging, remote sensing and road signs. In the following experiments, some images issued from the above-mentioned applications are used.
For the X-ray hand image, provided in Fig.~\ref{Hand_Xray}, our proposed method aims to detect the five linear structures formed by the  phalanges and metacarpals bones, corresponding to the five hand fingers. As first remark, we observe that, visually, the centerlines of the bone structures corresponding to the thumb and the little finger, estimated by Algo 2 using Hessian-based angle initialization (see Fig.~\ref{Hand_Xray}b), seem to be more accurate than the centerlines estimated by Algo 1 using the same initialization (see Fig.~\ref{Hand_Xray}a). In fact, the centerlines obtained by Algo 2, compared to those obtained by Algo 1, are closer to the axes of symmetry of the real linear structures fitted by the proposed linear anchored Gaussian mixture model. The outstanding accuracy obtained with Algo 2 is obtained thanks to its design, where the part situated between the thumb and the index finger is considered as irrelevant and then, is ignored during the computation of the likelihood function. 
The same observations are made on the satellite image, the road lane marking and the T-weld radiographic image, where the best results are obtained by Algo 2 using Hessian-based angle initialization, where the linear structures centerlines and thicknesses are accurately recovered as illustrated in Figs.~\ref{real_im_res}a,~\ref{real_im_res}b and \ref{real_im_res}c, respectively.  
 
\section{Conclusion}
In this paper, a novel paradigm to better detect thick linear objects is presented. Compared to all the material found in the literature related to this domain, our method stands out for the simplicity of its implementation, where the linear structure gray level 3D representation is modeled as linear anchored Gaussian, characterized by scale and location parameters. The estimation of the latter are done in the framework of finite mixture model using EM algorithm as parameter estimator. Satisfactory results are obtained on noise/blur-free and noisy/blurry synthetic images, where the linear structures location and size are accurately recovered. However, with the presence of strong noise which affects also the background, some errors can occur on the detection of linear structures with small sizes or near the boundaries of the image. By computing the image Hessian, the first improvements of the proposed method come to make an automatic selection of the mixture model and to get an accurate initial angle parameter for EM algorithm. Then, inserting a background subtraction step at each EM algorithm iteration permits to rid the data, involved in the likelihood function computation, of irrelevant information brought by nonuniform and noisy background. A the end of this stage, as confirmed by the used performance measures, the new version of our method achieves very accurate thickness measure and centerline location of the synthetic images used in test even with strong blur and noise. Qualitatively judged, the results obtained for real images are very satisfying, where very subtle situations of thick object detection are successively achieved.      

\appendix
\section{Geometric configurations after sliding $z$-axis perpendicularly to $\Delta$ in Fig.~\ref{l_Gauss}}\label{apdx_A}
Depending on the position of $z$-axis origin on the $x-y$ axes system, we can distinguish 3 configurations (see Fig.~\ref{l_Gauss}): (1) $\lbrack Az$, (2) $\lbrack Bz$, and (3) $\lbrack Oz$. 

\begin{enumerate}
\item\label{item_1} $\lbrack Az$ axis: Assume $(a,0)$ the $x-y$ coordinates of $A$. The values of the data element $z$ and the mean parameter $\mu_z$, represented in $x-y$ coordinates are given by $z=\overline{AP_z}=(x-a)\cos\theta+y\sin\theta$ and $\mu_z=\overline{AP_\mu}=\sqrt{(\mu_x-a)^2+\mu_y^2}$, respectively. Using $y=x\tan\theta-a\tan\theta$ as the equation of $z$-axis in $x-y$ plane and knowing that $P_\mu$ is the intersection of $\lbrack Az$ and $(\Delta)$, we obtain the following system of equation for the mean parameter:
$\{\mu_x\tan\theta-\mu_y = a\tan\theta\ ;\ \mu_x\cos\theta+\mu_y\sin\theta=\rho\}$ of which solutions are $\mu_x=\frac{\rho/\sin\theta+a\tan\theta}{\tan\theta+\cot\theta}$ and $\mu_y= \left[\frac{\rho/\sin\theta+a\tan\theta}{\tan\theta+\cot\theta}-a\right]\tan\theta$.   
The mean parameter along $z$-axis is deduced as $\mu_z=\rho-a\cos\theta$. 

\item $\lbrack Bz$ axis: Assume $(0,b)$ the $x-y$ coordinates of the $z$-axis origin $B$. By following the same reasoning as in Item~\ref{item_1}, we obtain $\mu_x=\frac{\rho/\sin\theta-b}{\tan\theta+\cot\theta}$ and $\mu_y= \frac{\rho/\sin\theta-b}{\tan\theta+\cot\theta}\tan\theta+b$. The mean parameter along $z$-axis is deduced as $\mu_z=\sqrt{\mu_x^2+(\mu_y-b)^2}=\rho-b\sin\theta$.

\item $\lbrack Oz$ axis: This case is obtained when $a$ or $b$ are null. The values of the data element $z$ and the mean parameter $\mu_z$ are then deduced as $z=\overline{OP_z}=x\cos\theta+y\sin\theta$ and $\mu_z=\rho$, respectively.
\end{enumerate}

\section{Scale and thickness relationship for a linear structure}\label{apdx_B}
\begin{figure}[tbh!]
\centering
\includegraphics[scale=0.6]{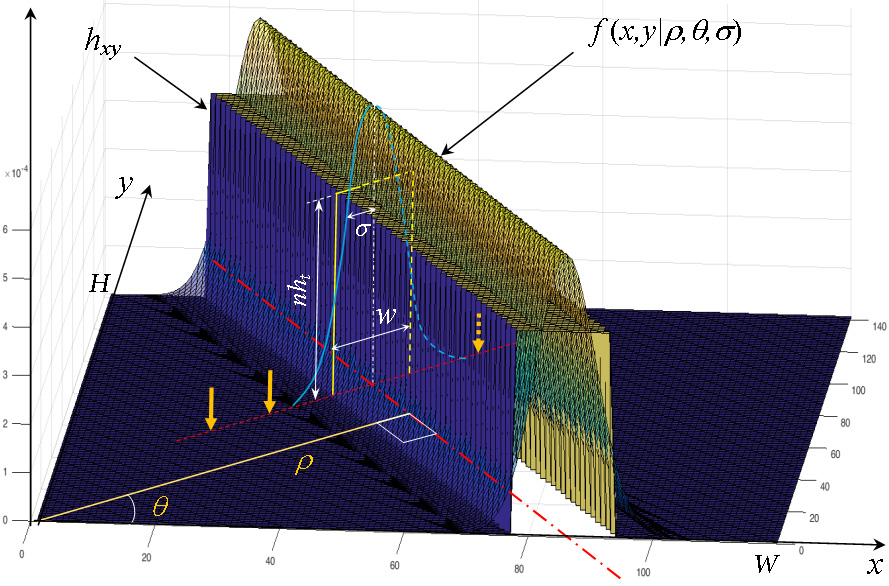}
\caption{Thick line fitted by directional Gaussian distribution.}
\label{bar_fit}
\end{figure}

\begin{figure}[tbh!]
\centering
\includegraphics[scale=0.6]{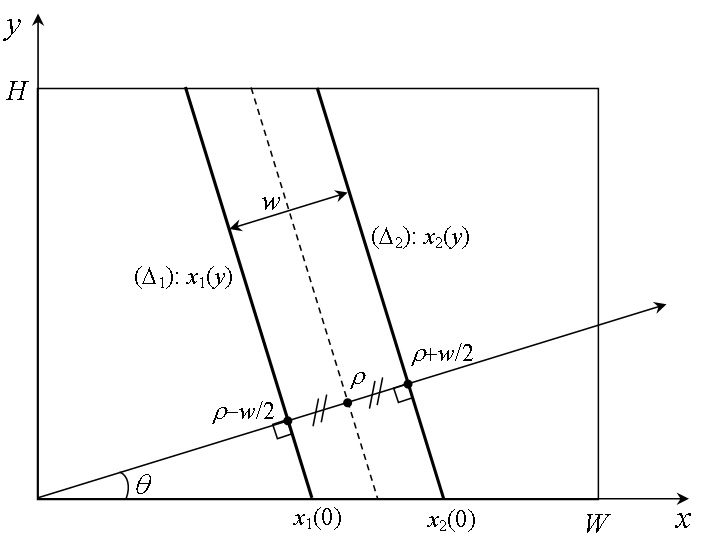}
\caption{Thick line projections on $x-y$ plane.}
\label{bar_proj}
\end{figure}

Let examine, in case of an image presenting a perfect thick line of width $w$, as illustrated in Fig.~\ref{bar_fit}, the relationship between the line width and the scale parameter $\sigma$ which is nothing but the standard deviation of the directional Gaussian distribution fitting the thick line in question. 
In this case, referring to Fig.~\ref{bar_fit}, the following parameters, used in Eq.~\ref{sig_est}, are set as follows: $M=1$ (since there is a unique object the subscript $m$ is dropped), $\pi_m=1$, $z_{xym}=\mathbb{1}_{W\times H}$ and $h_{xy}=nh_t=\frac{h_t}{\sum_x\sum_y I_{xy}}$ which represents the normalized gray level value of the thick line. Using the notations given in Fig.~\ref{bar_proj}, illustrating the projection of a thick line relief on $x-y$ plane, the following line equations are derived    
\setlength{\abovedisplayskip}{4pt}
\setlength{\belowdisplayskip}{4pt}
\begin{align}
(\Delta_1) : x = x_1(y)=\frac{\rho-w/2}{\cos\theta}-y\tan\theta\\
(\Delta_2) : x = x_2(y)=\frac{\rho+w/2}{\cos\theta}-y\tan\theta
\end{align}
Considering that $h_{xy}=nh_t$, for all the pixels inside the space domain, delimited by the lines $(\Delta_1)$ and $(\Delta_2)$ and taking a double integration on simple domain instead of double summation, Eq.~\ref{sig_est} becomes
\begin{align}\nonumber
\sigma^2 =& nh_t\int\limits_{y=0}^{y=H}\left[\int\limits_{x=x_1(y)}^{x=x_2(y)}(x\cos\theta+y\sin\theta-\rho)^2\dd x\right]\dd y\\
=&nh_t\int\limits_{0}^{H}\left[\int\limits_{\frac{\rho-w/2}{\cos\theta}-y\tan\theta}^{\frac{\rho+w/2}{\cos\theta}-y\tan\theta}(x\cos\theta+y\sin\theta-\rho)^2\dd x\right]\dd y = nh_t\frac{H w^3}{12\cos\theta}
\label{sig_width_rel}
\end{align}

In this configuration, the thick line volume $v_l$ can be computed as $\frac{h_tHw}{\cos\theta}$ and, consequently, $nh_t=h_t/v_l=\frac{\cos\theta}{Hw}$. Then, from Eq.~\ref{sig_width_rel}, we obtain
\setlength{\abovedisplayskip}{4pt}
\setlength{\belowdisplayskip}{4pt}
\begin{equation}
\sigma^2 = \frac{\cos\theta}{Hw}\times\frac{H w^3}{12\cos\theta}=\frac{w^2}{12}
\end{equation}

Finally, the relationship linking the standard deviation of a directional Gaussian fitting a thick line to the width of the latter is given by
\begin{equation}
2\sigma=w/\sqrt{3}
\end{equation}

\bibliographystyle{unsrt}
\bibliography{mybibfile}


\end{document}